\documentclass[11pt]{article}

\usepackage[letterpaper,margin=1in]{geometry}  
\usepackage{times}                              
\usepackage[round]{natbib}                      
\usepackage[table]{xcolor}                      

\usepackage{amsmath,amsfonts,bm}

\def\eqref#1{equation~\ref{#1}}

\def\1{\bm{1}}
\newcommand{\train}{\mathcal{D}}

\DeclareMathAlphabet{\mathsfit}{\encodingdefault}{\sfdefault}{m}{sl}
\SetMathAlphabet{\mathsfit}{bold}{\encodingdefault}{\sfdefault}{bx}{n}

\DeclareMathOperator*{\argmax}{arg\,max}

\usepackage[utf8]{inputenc}
\usepackage[T1]{fontenc}
\usepackage{microtype}
\usepackage{graphicx}
\usepackage{subcaption}
\usepackage{booktabs}

\usepackage{amssymb}
\usepackage{mathtools}
\usepackage{amsthm}
\usepackage{nicefrac}

\usepackage{bbm}          
\usepackage{enumitem}     
\usepackage{algorithm}    
\usepackage{algpseudocode}
\usepackage{xcolor}
\usepackage{multirow}

\usepackage{wrapfig}
\usepackage{caption}
\usepackage{tikz}
\usepackage{pgfplots}
\pgfplotsset{compat=1.18}
\usetikzlibrary{positioning,arrows.meta,shapes.geometric,fit,backgrounds,calc,patterns,decorations.pathreplacing}

\theoremstyle{plain}
\newtheorem{theorem}{Theorem}[section]
\newtheorem{proposition}[theorem]{Proposition}

\theoremstyle{definition}

\theoremstyle{remark}
\newtheorem{remark}[theorem]{Remark}

\newcommand{\Pup}{\overline{P}}
\newcommand{\Plow}{\underline{P}}
\newcommand{\regen}[1]{\textcolor{red}{#1}}
\newcommand{\Dset}{\mathcal{D}}
\newcommand{\Hyp}{\mathcal{H}}
\newcommand{\Frame}{\mathcal{C}}

\usepackage{url}
\usepackage{hyperref}
\usepackage[capitalize,noabbrev]{cleveref}
\crefname{equation}{Eq.}{Eqs.}
\Crefname{equation}{Eq.}{Eqs.}
\crefname{section}{Sec.}{Secs.}
\Crefname{section}{Sec.}{Secs.}
\crefname{subsection}{Sec.}{Secs.}
\Crefname{subsection}{Sec.}{Secs.}
\crefname{subsubsection}{Sec.}{Secs.}
\Crefname{subsubsection}{Sec.}{Secs.}
\crefname{appendix}{\S}{\S}
\Crefname{appendix}{\S}{\S}
\crefname{table}{Tab.}{Tabs.}
\Crefname{table}{Tab.}{Tabs.}
\crefname{figure}{Fig.}{Figs.}
\Crefname{figure}{Fig.}{Figs.}

\title{Credal Large Language Models \\
  for Semantic Commitment under Uncertainty}

\makeatletter
\newcommand{\AND}{\end{tabular}\par\vspace{1em}\begin{tabular}[t]{c}}
\makeatother

\author{%
  \textbf{Shireen Kudukkil Manchingal} \\
  Oxford Dynamics \\
  Oxford \\
  \texttt{smanchingal@brookes.ac.uk}
  \and
  \textbf{Sofiia Nikolenko} \\
  Ludwig-Maximilians-Universit\"at M\"unchen \\
  Munich \\
  relAI -- Konrad Zuse School of \\
  Excellence in Reliable AI \\
  \texttt{sofiia.nikolenko@campus.lmu.de}
  \AND
  \textbf{Fabio Cuzzolin} \\
  Institute for Artificial Intelligence, Data Analysis and Systems (AIDAS) \\
  School of Engineering Computing \& Mathematics \\
  Oxford Brookes University, Oxford, UK \\
  \texttt{fabio.cuzzolin@brookes.ac.uk}
}
\date{}

\begin{document}

\maketitle
\renewcommand\thefootnote{}


\begin{abstract}
Large language models (LLMs) often produce fluent but incorrect answers with unwarranted confidence. A central limitation is that standard LLMs represent uncertainty through a single predictive distribution, conflating epistemic ignorance with genuine ambiguity. We introduce Credal Large Language Models (CLLMs): an ensemble of LoRA adapters induces a credal set whose lower and upper probabilities expose the spread of plausible predictive distributions rather than collapsing to a single softmax output. From this representation, we derive a single commitment rule: the model commits to an answer only when its lower probability exceeds the upper probability of every alternative, and otherwise returns the set of answers that no plausible predictor rules out. We apply this commitment rule at two depths: Credal Token Commitment (CTC) applies it to answer tokens from one ensemble forward pass, which decides constrained answers without any generation; for open-ended answers, credal decoding extends a partial answer only when no completed answer dominates it, so that the completions produced are those the plausible predictors license, and Credal Semantic Commitment (CSC) applies the rule to their meaning clusters. We evaluate CLLMs with Gemma-2-9B, Llama-3.1-8B and Qwen2.5-7B on OpenBookQA, CoQA, TriviaQA and ARC-Challenge. On multiple choice, CTC commits on 73--91\% of questions at 89--98\% accuracy, returns sets of 1.1--1.5 options containing the gold one on 89--98\%, and its intervals contain the observed accuracy in 24 of 30 confidence bins without calibration; corrupted context lowers commitment from 87--92\% to 65--71\%, and on Gemma the credal bound detects corruption better than every baseline. On open-ended QA, CLLM outperforms semantic entropy and Laplace-LoRA at a fixed coverage by up to 19\% and 9.5\% absolute accuracy on CoQA and TriviaQA with context, for every backbone.
\end{abstract}


\section{Introduction}
\label{sec:intro}
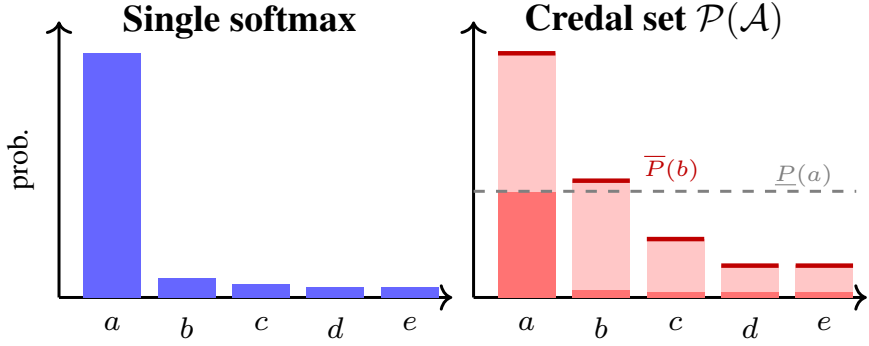
\begin{figure}
    \centering
    \resizebox{0.7\linewidth}{!}{%
    \begin{tikzpicture}[every node/.style={font=\small}]
        \begin{scope}[xshift=0cm]
        \node[anchor=south,font=\small\bfseries] at (1.7,2.3) {Single softmax};
        \draw[->,thick] (0,0) -- (0,2.6);
        \draw[->,thick] (0,0) -- (3.7,0);
        \node[rotate=90,anchor=south,font=\scriptsize] at (-0.1,1.3) {prob.};
        \foreach \x/\h/\lab in {0.5/2.3/$a$, 1.2/0.18/$b$, 1.9/0.13/$c$, 2.6/0.10/$d$, 3.3/0.10/$e$} {
            \fill[blue!60] (\x-0.27,0) rectangle (\x+0.27,\h);
            \node[font=\scriptsize,anchor=north] at (\x,-0.05) {\lab};
        }
        \end{scope}
        \begin{scope}[xshift=3.9cm]
        \node[anchor=south,font=\small\bfseries] at (1.7,2.3) {Credal set $\mathcal{P}(\mathcal{A})$};
        \draw[->,thick] (0,0) -- (0,2.6);
        \draw[->,thick] (0,0) -- (3.7,0);
        \foreach \x/\hl/\hu/\lab in {0.5/1.0/2.3/$a$, 1.2/0.07/1.1/$b$, 1.9/0.06/0.55/$c$, 2.6/0.05/0.30/$d$, 3.3/0.05/0.30/$e$} {
            \fill[red!55] (\x-0.27,0) rectangle (\x+0.27,\hl);
            \fill[red!22] (\x-0.27,\hl) rectangle (\x+0.27,\hu);
            \draw[red!75!black,very thick] (\x-0.27,\hu) -- (\x+0.27,\hu);
            \node[font=\scriptsize,anchor=north] at (\x,-0.05) {\lab};
        }
        \draw[dashed,gray,thick] (0,1.0) -- (3.6,1.0);
        \node[font=\tiny,gray,anchor=west] at (2.7,1.15) {$\Plow(a)$};
        \node[font=\tiny,red!75!black,anchor=west] at (1.47,1.22) {$\Pup(b)$};
        \end{scope}
    \end{tikzpicture}%
    }%
    \caption{\textbf{Left:} one softmax over {five candidate answers $a$-$e$} is sharp whether the model knows the answer or merely settled on one. \textbf{Right:} the credal set {$\mathcal{P}(\mathcal{A})$ (\cref{eq:credalset})} of five LoRA adapters gives each answer a lower (solid) and an upper (band) probability. Here the lower bound of $a$ (dashed) is below the upper bound of $b$, so some adapter prefers $b$: the model should not commit to $a$ alone.}
    \label{fig:motivation}
\end{figure}

Large Language Models (LLMs) achieve strong performance on question answering, reasoning, code generation and open-ended dialogue~\citep{brown2020language, chowdhery2023palm, touvron2023llama, jiang2023mistral7b}, yet they continue to produce fluent-but-incorrect answers with unwarranted confidence, particularly when context is incomplete, conflicting or adversarial~\citep{maynez2020faithfulness, ji2023survey, huang2023survey, farquhar2024detecting}. 
In safety-critical settings the model often fails not by being uncertain, but by appearing \emph{insufficiently uncertain}. 
A deployed system needs to make a decision for each question: whether the model may answer and, if not, which answers remain plausible. A standard LLM cannot supply one, because it represents its uncertainty through a single next-token distribution obtained by softmax normalisation (\cref{fig:motivation}). One such distribution can be sharp for two very different reasons: because the evidence supports one answer, or because fine-tuning happened to settle on one of several answers that the data does not adjudicate between.

Existing methods build a family of models and then summarise it. Bayesian and Laplace approximations place distributions over adapter parameters~\citep{yang2023bayesian, wang2024blob}, and LoRA~\citep{hu2021lora} ensembles, the most widely used remedy at LLM scale, fine-tune several adapters on the same data and average their next-token distributions~\citep{lakshminarayanan2017simple, balabanov2024uncertainty, wang2023loraensembles}. 
Averaging is where the information is lost. Five adapters that each prefer a different answer and five that agree on a hedged answer can give the same mean, so the pooled model reports how spread out its distribution is, not whether its members agree~\citep{hullermeier2021aleatoric}, and the disagreement survives only as a scalar, such as mutual information, that must again be thresholded. 
Generation compounds the loss: an open-ended answer has no fixed label set, the adapters often disagree about how an answer begins (\emph{Special} or \emph{special}) while agreeing on what it says. Sampling from the mean, or from one model as semantic entropy does~\citep{kuhn2023semantic, farquhar2024detecting}, yields candidates without any adapter's probability of them. Credal readings of ensembles in classification keep the members as a set of distributions and improve epistemic uncertainty over averaging~\citep{shaker2021ensemble, wang2024credalwrapper, wang2024credal}, as do set-valued classifiers such as random-set neural networks~\citep{manchingal2025rsnn}; however, they assume a fixed label set and return another score. In every case the family is collapsed to one distribution or one scalar before a decision is taken, and none of these methods tells a deployed system when it is in a position to answer and when it must present alternatives.

\subsection{Credal Large Language Models}

We address this from the perspective of \emph{imprecise probability}~\citep{walley1991statistical, levi1980enterprise}, in which epistemic uncertainty is represented by a \emph{set} of plausible distributions rather than a single point. We propose \textbf{Credal Large Language Models} (CLLMs): an ensemble of LoRA adapters on a frozen backbone induces a credal set, a closed convex set of distributions that bounds each candidate answer's probability from below and above, and the model \emph{commits} to a candidate only when no other candidate could be preferred by any distribution in that set. \textit{This is the interval-dominance rule of imprecise-probability decision theory~\citep{troffaes07}, and it changes the question from ``how uncertain is the prediction?'' to \textbf{``which answers does the evidence still license?''}.} The rule contains no threshold, temperature or sharpness parameter, reduces to argmax prediction when the adapters agree, and returns the set of licensed answers; the model commits when exactly one remains.


\subsection{Contributions}

Our main {contributions} are: \textbf{(i)} \emph{\textbf{CLLMs}}, which read a fine-tuned LoRA ensemble on a frozen backbone as a credal set, bound each candidate answer's probability by its minimum and maximum over the adapters, and decide by interval dominance, committing when one answer survives and otherwise returning the answers the evidence still licenses (\S\ref{sec:method:bounds}--\S\ref{sec:method:rule}); \textbf{(ii)} \textbf{\emph{Credal Token Commitment (CTC)}}, the interval-dominance rule of (i) applied to the whole vocabulary at the first answer position after one ensemble forward pass, committing when a single token survives, which decides multiple-choice questions with no generation (\S\ref{sec:method:depths}); \textbf{(iii)} \textbf{\emph{credal decoding}}, which extends a partial answer only while its largest probability under any adapter is at least the best completed answer's smallest probability over adapters, records every adapter's probability of each completion, and adds to the upper bounds the unexplored mass, the probability each adapter gives to completions never written out, so it needs no temperature or sample count and unwritten completions can only make the model more cautious (\S\ref{sec:method:decoding}); \textbf{(iv)} \textbf{\emph{Credal Semantic Commitment (CSC)}}, which groups completed answers into \textit{meanings} that every adapter judges equivalent, bounds each meaning by the adapters' summed probabilities, adding the unexplored mass to the upper bound, and applies the same interval dominance rule, so the model commits to what an answer says rather than to how it is spelled (\S\ref{sec:method:depths}); and \textbf{(v)} an evaluation of selective accuracy, calibration, set-valued prediction, hallucination and corrupted-context detection, and abstention on five datasets, three backbones and six baseline families, showing that the rule commits on the questions the model knows and withdraws when the evidence is corrupted, with no threshold.

\subsection{Paper outline}

\S\ref{sec:related} reviews related work; \S\ref{sec:method} defines the credal set and its bounds (\S\ref{sec:method:bounds}), the commitment rule (\S\ref{sec:method:rule}), credal decoding (\S\ref{sec:method:decoding}) and CTC and CSC (\S\ref{sec:method:depths}); \S\ref{sec:experiments} gives the setup (\S\ref{sec:exp:setup}), results (\S\ref{sec:exp:results}) and limitations (\S\ref{sec:exp:limitations}); and \S\ref{sec:conclusion} concludes. The appendix gives proofs (\S\ref{app:proofs}), extended related work (\S\ref{app:extended_related_work}), the algorithm with an example (\S\ref{app:methodology}), implementation, baselines and setup (\S\ref{app:implementation}), additional results on selective accuracy and calibration (\S\ref{app:results:comparison}), detection (\S\ref{app:results:detection}), abstention (\S\ref{app:results:selective}), set-valued prediction (\S\ref{app:results:sets}), the four-option setting (\S\ref{app:results:mcq}), the two depths (\S\ref{app:results:depths}), ablations (\S\ref{app:ablations}), cost-sensitive commitment (\S\ref{app:ablation:cost_sensitive}) and computational cost (\S\ref{app:cost}), and broader impact (\S\ref{app:impact}).

\section{Related Work}
\label{sec:related}

\subsection{Second-order uncertainty representations}

\emph{Imprecise probability} \cite{augustin2014introduction,beer2013imprecise,bradley2019imprecise,coolen2025imprecise} and robust Bayesian analysis represent epistemic uncertainty by a set of plausible distributions \cite{cuzzolin2021uncertainty,cuzzolin2025uncertainty} rather than a selected mixture~\citep{walley1991statistical, levi1980enterprise, berger1994robust}, with belief functions \citep{shafer1976mathematical,cuzzolin2014belief,cuzzolin2018visions,cuzzolin2020geometry,cuzzolin2026statistical} and credal uncertainty measures as related tools~\citep{cuzzolin2014belief, cuzzolin2024uncertainty, gonzalezgarcia_2026_credal}. 
A geometric approach to belief functions and other uncertainty measures can be found in \cite{cuzzolin2004geometry,cuzzolin2008geometric,cuzzolin2008credal,cuzzolin2020geometry-dempster}; their combinatorial analysis in \cite{cuzzolin2010three}, their algebraic properties in \cite{cuzzolin01bcc,cuzzolin2001lattice,zhou2017total}.

Credal-set and random-set neural networks bring these representations to classification~\citep{wang2024creinns, wang2024credalwrapper, manchingal2025rsnn}, and recent work carries credal ideas to language, through credal sets of creative continuations~\citep{garcesarias_2025_geometry} and attention parameterised by Dirichlet evidence~\citep{ji_2025_credal}. These works assume a fixed label set, measure diversity rather than choose an answer. To our knowledge, CLLMs are the first to use interval-dominance decisions to decide whether an LLM should answer, in both next-token prediction and open-ended generation; an extended discussion is in \S\ref{app:extended_related_work}.

\emph{Second-order probability representations} have been applied in the past to pose estimation (\cite{gong2017belief}), tracking (\cite{cuzzolin1999evidential}), logistic regression (\cite{cuzzolin2018belief}), clustering (\cite{zhu2026tdcc}), feature fusion (\cite{pang2025guest}), max-entropy classification (\cite{cuzzolin2018generalised}), statistical learning theory (\cite{cuzzolin2024generalising,caprio2024credal}) and a variety of other tasks (\cite{liu2019evidence,faza2026direct,manchingal2025uncertainty}).
A recent review can be found in \cite{wang2025review}.

Within imprecise-probabilistic, second-order uncertainty methods applied to machine learning, two prominent approached are based on random sets and credal sets, respectively.

\subsubsection{Random-set models} 

\emph{Random-set} models,
within the approach termed `epistemic AI' (\cite{cuzzolin2024epistemic,manchingal2024epistemic,manchingal2025epistemic}), allow probability to be assigned to sets of possible
outcomes without specifying how it is divided among their members
\citep{shafer2020mathematical,cuzzolin2020geometry}.
Random-Set Neural Networks (RS-NN) implement this representation for
classification, including text classification
\citep{manchingal2025rsnn}.
Random-Set Large Language Models (RS-LLM) extend it to language generation,
predicting over sets of possible next tokens
\citep{mubashar2025rsllm}.
Other approaches obtain lower and upper probabilities without predicting
probabilities for answer sets, for instance those using credal sets, i.e., convex sets of distributions \cite{levi1980enterprise,caprio2025credal}. 

\subsubsection{Credal-set models} 

Within \emph{credal set} representations, Credal Deep Ensembles combine networks
trained to predict probability bounds \citep{wang2024credalensembles,wang2026learning,wang2026credal},
while the Credal Wrapper constructs bounds from the predictions of
Bayesian neural networks or deep ensembles
\citep{wang2025credalwrapper}, using the intersection probability (\cite{cuzzolin2009credal,cuzzolin2007properties}) as pointwise estimator. Credal INNs \cite{wang2025creinns} combine the expressive power of interval neural networks with credal representations.
For language models, \citet{yang2026imprecise} investigate prompting
models to report lower and upper probabilities for possible answers,
as well as constructing intervals from multiple reported predictions.
The credal approach has been recently extended to generative adversarial networks (\cite{mubashar2026epistemic}) and to a wrapper applicable to  any Bayesian NN (\cite{sultana2025epistemic}).

\subsection{Uncertainty in Large Language Models}

\emph{Uncertainty in LLMs} is most often quantified by a scalar read off the predictive distribution. The aleatoric\slash epistemic split is formalised at the level of that distribution~\citep{hullermeier2021aleatoric}, although linguistic variation complicates it for generated text~\citep{baan_2023_uncertainty, bickfordsmith_2025_rethinking}. LLMs are systematically miscalibrated on question answering and few-shot tasks~\citep{jiang2021how, zhao2021calibrate}; sequence-level scores can be obtained by ensembling and decomposed into token-level contributions~\citep{malinin2021uncertainty}, verbalised confidence is often a better signal than the probabilities of RLHF-tuned models~\citep{tian2023just, lin2022teaching, kadavath2022language}, and semantic calibration can be improved after the fact~\citep{lamb_2026_temperature, dixon_2026_semantic}. Conformal Language Modeling calibrates sampling and rejection to return answer sets with coverage guarantees~\citep{quach2024conformal}, and later conformal methods drop the need for logits, trade specificity for factuality or learn when to abstain~\citep{su_2024_api, mohri_2024_factuality, jiang_2025_linguistic, wang_2025_sconu, tayebati_2025_cap}. In every case the model answers when a calibrated threshold on a scalar is met, whereas CLLMs answer only when the evidence licenses a single answer, with no fitted cutoff.
Approximate posteriors retain more than a single distribution.

Full Bayesian inference is intractable at LLM scale, so recent work places the posterior over the parameters of LoRA adapters~\citep{hu2021lora}: Laplace-LoRA fits a KFAC Gaussian posterior around the fine-tuned weights~\citep{yang2023bayesian}, BLoB trains a variational posterior through an ELBO~\citep{wang2024blob}, and LoRA ensembles replace the explicit posterior with several independently trained adapters~\citep{balabanov2024uncertainty,wang2023loraensembles}.
Related methods combine expert adapters, learn calibrated mixtures of LoRA experts or perturb low-rank weights during decoding~\citep{li_2025_experts, niu2024uq4ct, zhang2025tokur}. Each constructs a family of predictors and then summarises it, as a mean distribution, a mixture or a disagreement score, before any decision is taken; CLLMs keep the family as a set, so whether its members agree becomes the decision itself.

\subsection{Hallucination and abstention}

\emph{Hallucination detection and abstention} are instances of selective prediction with a reject option~\citep{huang2023survey, ji2023survey}. Because many surface forms express one answer, token-level dispersion is a poor hallucination signal; semantic entropy instead clusters sampled generations by bidirectional entailment and scores the diversity of the clusters~\citep{kuhn2023semantic, farquhar2024detecting}. Later work keeps graded similarities or semantic distances instead of hard classes~\citep{nikitin_2024_kernel, qiu_2024_density, lin_2024_confidence}, measures consistency across sampled answers~\citep{wang2023selfconsistency, manakul_2023_selfcheck}, approximates it from hidden states or corrects it for incomplete sampling~\citep{kossen_2024_probes, mccabe_2026_alphabet}, or varies the input or the model~\citep{hou2024decomposing, li_2025_esi, li_2026_volume, hamidieh_2026_disagreement}. Most of these reason about the frequencies of samples from one model. 

Credal decoding, in opposition, writes out the candidates a set of predictors licenses, obtains each predictor's probability of them and accounts for the probability each predictor gives to completions it never wrote out, so the candidates, their probabilities and their grouping into meanings all come from the same predictors, rather than from one model's samples and an external entailment model, as in semantic entropy.

\section{Methodology}
\label{sec:method}

\begin{figure}[!h]
\vspace{-8pt}
\centering
\providecommand{\sym}[2]{{\itshape #1}\textsubscript{#2}}
\resizebox{\textwidth}{!}{%
\begin{tikzpicture}[
  x=1cm,y=1cm,
  every node/.style={font=\fontsize{6.7}{7.5}\selectfont,inner sep=0pt},
  heading/.style={font=\fontsize{7.6}{8.1}\selectfont\bfseries},
  small/.style={font=\fontsize{6.0}{6.7}\selectfont},
  micro/.style={font=\fontsize{5.7}{6.3}\selectfont},
  box/.style={draw,line width=.45pt,rounded corners=3pt,align=center,
    inner xsep=3pt,inner ysep=2pt},
  ada/.style={box,draw=olive!60!black,fill=olive!15,
    minimum width=1.27cm,minimum height=.37cm},
  completion/.style={box,draw=orange!70!black,fill=orange!12,
    font=\fontsize{6.0}{6.6}\selectfont,minimum width=3.00cm,minimum height=.30cm},
  arr/.style={-{Latex[length=1.35mm,width=.9mm]},line width=.65pt},
  bound/.style={line width=.9pt},
  cutoff/.style={red!70!black,densely dashed,line width=.55pt}
]
\colorlet{tok}{blue!70!black}
\colorlet{sem}{orange!90!black}
\colorlet{rule}{red!75!black}
\colorlet{commit}{green!55!black}
\path[use as bounding box] (0,-.04) rectangle (17.25,5.45);

\draw[blue!70,draw opacity=.35,line width=.8pt,rounded corners=9pt]
  (3.63,2.78) rectangle (11.35,5.1);
\draw[orange!80!black,draw opacity=.35,line width=.8pt,rounded corners=9pt]
  (3.63,.00) rectangle (11.35,2.70);
\filldraw[fill=red!4,draw=rule,rounded corners=3pt,line width=.45pt]
  (11.67,2.82) rectangle (15.30,4.90);
\filldraw[fill=red!4,draw=rule,rounded corners=3pt,line width=.45pt]
  (11.67,.00) rectangle (15.30,2.62);

\node[box,draw=black!80,fill=blue!8,minimum width=1.20cm,
  minimum height=.65cm] (input) at (.64,2.60) {Input\\prompt {\itshape x}};
\node[small,align=center,text width=1.45cm] at (.71,1.91)
  {Where does the\\train terminate?};
\node[box,draw=olive!60!black,fill=olive!8,minimum width=1.82cm,
  minimum height=3.45cm] (llm) at (2.36,2.60) {};
\node[heading,text=olive!60!black] at (2.36,4.62) {(1) CLLM};
\node[font=\fontsize{7.2}{8}\selectfont\bfseries] at (2.36,4.01) {LLM};
\node[small] at (2.36,3.71) {(frozen backbone)};
\node[ada] at (2.36,3.16) {LoRA \sym{f}{1}};
\node[ada] at (2.36,2.59) {LoRA \sym{f}{2}};
\node at (2.36,2.05) {\ldots};
\node[ada] at (2.36,1.48) {LoRA \sym{f}{M}};
\draw[arr] (input.east)--(llm.west);
\draw[arr] ([yshift=1.15cm]llm.east)--(3.60,3.75);
\draw[arr] ([yshift=-1.39cm]llm.east)--(3.88,1.21);

\node[heading,text=blue!80] at (7.49,4.9) {(2a) Token space};
\begin{scope}[shift={(4.03,3.66)}]
  \draw[-{Latex[length=.8mm]},line width=.35pt] (-.03,0)--(1.35,0);
  \draw[-{Latex[length=.8mm]},line width=.35pt] (0,-.03)--(0,.94);
  \foreach \x/\h in {.14/.768,.60/.329,1.06/.274}
    \fill[blue!55] (\x,0) rectangle +(0.18,\h);
  \node[micro,anchor=north] at (.23,-.08) {Paris};
  \node[micro,anchor=north] at (.69,-.08) {The};
  \node[micro,anchor=north] at (1.15,-.08) {Lyon};
  \node[small,text=blue!50!black] at (.66,1.09) {\sym{p}{m}(\textperiodcentered\,|\,{\itshape x})};
\end{scope}
\node[small,text=blue!50!black] at (4.62,2.98) {{\itshape M} distributions};
\draw[arr] (5.42,3.90)--(5.64,3.90);
\begin{scope}[shift={(5.68,3.31)},scale=.68]
  \coordinate (A) at (0,0); \coordinate (B) at (2,0); \coordinate (C) at (1,1.73205);
  \fill[blue!4] (A)--(B)--(C)--cycle;
  \draw[blue!50,line width=.65pt] (A)--(B)--(C)--cycle;
  \foreach \q/\r in {.62/.13,.24/.2,.54/.21,.35/.09,.48/.14}
    \fill[blue!70!black] ({2*\q+\r},{1.73205*\r}) circle (.95pt);
  \node[small,anchor=north east,text=blue!60!black] at (.08,-.02) {Paris};
  \node[small,anchor=north west,text=blue!60!black] at (1.92,-.02) {The};
  \node[small,anchor=south,text=blue!60!black] at (1,1.75) {Lyon};
\end{scope}
\node[small,text=blue!70!black] at (6.36,2.98) {probability simplex};
\draw[arr] (7.08,3.90)--(7.34,3.90);
\begin{scope}[shift={(7.38,3.31)},scale=.68]
  \coordinate (A) at (0,0); \coordinate (B) at (2,0); \coordinate (C) at (1,1.73205);
  \fill[blue!4] (A)--(B)--(C)--cycle;
  \foreach \v/\a/\b/\lo/\hi in {A/B/C/.25/.56,B/A/C/.24/.62,C/A/B/.09/.21}{
    \draw[blue!45,densely dashed,line width=.38pt]
      ($(\a)!\lo!(\v)$)--($(\b)!\lo!(\v)$);
    \draw[blue!60,densely dashed,line width=.42pt]
      ($(\a)!\hi!(\v)$)--($(\b)!\hi!(\v)$);
  }
  \filldraw[fill=blue!13,draw=blue!45,densely dashed,line width=.45pt]
    (1.37,.225167)--(1.29,.363731)--(.69,.363731)--(.68,.346410)--(.79,.155885)--(1.33,.155885)--cycle;
  \filldraw[fill=cyan!45,fill opacity=.85,draw=blue!80!black,line width=.5pt]
    (.68,.346410)--(.79,.155885)--(1.37,.225167)--(1.29,.363731)--cycle;
  \draw[blue!50,line width=.65pt] (A)--(B)--(C)--cycle;
  \foreach \q/\r in {.62/.13,.24/.2,.54/.21,.35/.09,.48/.14}
    \fill[blue!70!black] ({2*\q+\r},{1.73205*\r}) circle (.95pt);
\end{scope}
\draw[blue!55,line width=.35pt] (8.10,3.04)--(8.16,3.44);
\node[small,text=blue!70!black] at (8.06,2.98) {credal set};
\draw[arr] (8.78,3.90)--(9.16,3.90);
\node[small,text=black!65,anchor=east] at (9.18,4.14) {min / max};
\node[box,draw=blue!60!black,fill=blue!25,minimum width=1.84cm,
  minimum height=.65cm] (tokenbounds) at (10.16,3.90) {Lower / upper\\probabilities};
\draw[arr] (tokenbounds.east)--(11.64,3.90);

\node[heading,text=sem] at (7.49,2.52) {(2b) Semantic space};
\node[small,align=center,text=sem] at (4.10,2.15) {Credal\\decoding};
\draw[orange!70!black,line width=.5pt]
  (3.94,1.21)--(4.13,1.70)--(4.40,1.82)
  (3.94,1.21)--(4.13,1.23)--(4.40,1.40)
  (4.13,1.23)--(4.40,.95)
  (3.94,1.21)--(4.13,.53)--(4.40,.53);
\foreach \x/\y in {3.94/1.21,4.13/1.70,4.13/1.23,4.13/.53}
  \draw[orange!70!black,fill=white,line width=.4pt] (\x,\y) circle (.8pt);
\foreach \y in {1.82,1.40,.95,.53}
  \fill[orange!70!black] (4.40,\y) circle (.85pt);
\draw[black!45,dashed,line width=.4pt] (4.13,.53)--(4.37,.26);
\draw[rule,line width=.6pt] (4.32,.21)--(4.42,.31) (4.32,.31)--(4.42,.21);
\node[micro,text=black!65,anchor=east] at (4.24,.26) {dominated};
\foreach \i/\yy/\sentence in {
  1/1.82/Paris is the final stop.,
  2/1.40/The train terminates in Paris.,
  3/.95/The train terminates in Lyon.,
  4/.53/Lyon is the final stop.}{
  \node[completion,anchor=west] (h\i) at (4.59,\yy)
    {\sym{h}{\i}: \sentence};
  \draw[arr,orange!70!black] (4.45,\yy)--(h\i.west);
}
\node[small,text=sem] at (8.46,2.15) {Group by meaning};
\filldraw[fill=orange!30,draw=orange!55,line width=.5pt]
  (8.47,1.64) ellipse (.59 and .40);
\filldraw[fill=orange!15,draw=orange!55,line width=.5pt]
  (8.47,.65) ellipse (.59 and .38);
\node[small] at (8.47,1.86) {Paris};
\node[small] at (8.47,.86) {Lyon};
\node[small,text=sem,anchor=west] at (9.02,1.90) {\sym{c}{1}};
\node[small,text=sem,anchor=west] at (9.02,.98) {\sym{c}{2}};
\foreach \i/\x/\y in {1/8.22/1.59,2/8.72/1.59,3/8.22/.59,4/8.72/.59}{
  \fill[orange!80!black] (\x,\y) circle (1.0pt);
  \node[micro,anchor=north] at (\x,\y-.07) {\sym{h}{\i}};
}
\draw[arr,orange!70!black] (h1.east)--(7.96,1.82);
\draw[arr,orange!70!black] (h2.east)--(7.99,1.40);
\draw[arr,orange!70!black] (h3.east)--(8.08,.90);
\draw[arr,orange!70!black] (h4.east)--(7.92,.53);
\node[micro,text=olive!60!black] at (6.47,.13) {merged only if every adapter agrees};
\draw[olive!60!black,densely dashed,line width=.4pt]
  (llm.south)--(2.36,.13)--(4.66,.13);
\draw[arr,olive!60!black,line width=.4pt] (4.66,.13)--(4.90,.13);

\draw[black!70,line width=.4pt]
  (9.06,1.64)--(9.22,1.64)--(9.22,.65)--(9.06,.65);
\draw[arr] (9.22,1.29)--(9.54,1.29);
\node[small,align=center] at (10.17,2.05) {Mass per\\adapter};
\foreach \i/\p/\q/\u in {
  1/.70/.12/.18,2/.66/.14/.20,3/.64/.135/.225,4/.55/.205/.245,5/.73/.085/.185}{
  \pgfmathsetmacro{\yy}{1.73-(\i-1)*.29}
  \node[micro,anchor=east] at (9.50,\yy) {\sym{f}{\i}};
  \fill[orange!60] (9.60,\yy-.065) rectangle ({9.60+1.03*\p},\yy+.065);
  \fill[orange!18] ({9.60+1.03*\p},\yy-.065) rectangle ({9.60+1.03*(\p+\q)},\yy+.065);
  \fill[black!3] ({9.60+1.03*(\p+\q)},\yy-.065) rectangle (10.63,\yy+.065);
  \fill[pattern=north east lines,pattern color=black!40]
    ({9.60+1.03*(\p+\q)},\yy-.065) rectangle (10.63,\yy+.065);
}
\fill[orange!60] (9.50,.22) rectangle (9.60,.32);
\node[micro,anchor=west] at (9.66,.27) {\sym{c}{1}};
\fill[orange!18] (9.97,.22) rectangle (10.07,.32);
\node[micro,anchor=west] at (10.13,.27) {\sym{c}{2}};
\fill[pattern=north east lines,pattern color=black!40] (10.45,.22) rectangle (10.55,.32);
\node[micro,anchor=west,text=black!65] at (10.61,.27) {\sym{u}{m}};
\draw[black!65,line width=.4pt] (10.66,1.82)--(10.82,1.82)--(10.82,.48)--(10.66,.48);
\draw[arr] (10.82,1.51)--(11.64,1.51);
\draw[black!60,line width=.4pt] (10.82,.48)--(11.00,.48)--(11.00,.67);
\draw[arr,black!60] (11.00,.67)--(11.64,.67);

\node[heading,text=rule] at (13.49,5.1) {(3) Commitment};
\node[font=\fontsize{6.8}{7.3}\selectfont,align=center] at (13.49,4.58)
  {Credal Token\\Commitment (\textbf{CTC})};
\foreach \yy/\lo/\hi/\lab/\col in {
  4.18/.25/.56/Paris/tok,3.89/.24/.62/The/tok,
  3.60/.09/.21/Lyon/black!40}{
  \node[small,anchor=east,text=\col] at (12.39,\yy) {\lab};
  \draw[bound,\col] ({12.53+1.15*\lo},\yy)--({12.53+1.15*\hi},\yy);
  \draw[\col,line width=.4pt]
    ({12.53+1.15*\lo},\yy-.05)--({12.53+1.15*\lo},\yy+.05)
    ({12.53+1.15*\hi},\yy-.05)--({12.53+1.15*\hi},\yy+.05);
}
\draw[cutoff] ({12.53+1.15*.25},3.44)--({12.53+1.15*.25},4.30);
\draw[black!45,line width=.5pt] (13.45,3.565)--(13.52,3.635)
    (13.45,3.635)--(13.52,3.565);
\draw[arr] (13.62,3.89)--(13.87,3.89);
\node[font=\fontsize{13}{13}\selectfont,text=rule] at (13.90,3.89) {\{};
\node[font=\fontsize{13}{13}\selectfont,text=rule] at (15.18,3.89) {\}};
\node[box,small,draw=tok,fill=blue!25,text width=.45cm,
  inner xsep=.9pt,minimum height=.28cm] at (14.28,3.89) {Paris};
\node[box,small,draw=tok,fill=blue!25,text width=.36cm,
  inner xsep=.9pt,minimum height=.28cm] at (14.86,3.89) {The};
\node[small,text=black!70] at (14.55,3.41) {two tokens};
\draw[cutoff] (12.29,3.20)--(12.52,3.20);
\node[micro,anchor=west,text=rule] at (12.61,3.20) {largest lower bound};
\node[box,draw=blue!55!black,fill=blue!8,minimum width=1.30cm,
  minimum height=.80cm] (tokenout) at (16.34,3.89) {No\\commitment};
\draw[arr] (15.33,3.89)--(tokenout.west);
\node[micro,text=black!62,align=center] at (16.34,3.34) {returns the set};

\node[font=\fontsize{6.8}{7.3}\selectfont,align=center] at (13.49,2.30)
  {Credal Semantic\\Commitment (\textbf{CSC})};
\node[small,anchor=east,text=sem] at (12.39,1.78) {\sym{c}{1}};
\node[small,anchor=east,text=black!40] at (12.39,1.38) {\sym{c}{2}};
\draw[bound,sem] ({12.53+1.15*.55},1.78)--({12.53+1.15*.73},1.78);
\draw[sem,dashed,line width=.65pt] ({12.53+1.15*.73},1.78)--({12.53+1.15*.915},1.78);
\draw[bound,black!40] ({12.53+1.15*.085},1.38)--({12.53+1.15*.205},1.38);
\draw[black!45,dashed,line width=.65pt] ({12.53+1.15*.205},1.38)--({12.53+1.15*.45},1.38);
\foreach \v/\yy/\col in {.55/1.78/sem,.915/1.78/sem,.085/1.38/black!40,.45/1.38/black!40}
  \draw[\col,line width=.4pt] ({12.53+1.15*\v},\yy-.05)--({12.53+1.15*\v},\yy+.05);
\draw[cutoff] ({12.53+1.15*.55},.54)--({12.53+1.15*.55},1.96);
\draw[black!45,line width=.5pt] (13.71,1.345)--(13.78,1.415) (13.71,1.415)--(13.78,1.345);
\draw[arr] (13.71,1.78)--(14.10,1.78);
\node[font=\fontsize{13}{13}\selectfont,text=rule] at (14.27,1.78) {\{};
\node[font=\fontsize{13}{13}\selectfont,text=rule] at (14.88,1.78) {\}};
\node[box,draw=orange!75!black,fill=orange!35,minimum width=.35cm,
  minimum height=.28cm] at (14.57,1.78) {\sym{c}{1}};
\node[small] at (14.57,1.29) {singleton};
\draw[arr] (14.57,1.59)--(14.57,1.45);
\draw[arr] (14.57,1.12)--(14.57,.97);
\draw[black!55,dashed,line width=.6pt] (12.53,.67)--({12.53+1.15*.245},.67);
\draw[black!55,line width=.4pt] ({12.53+1.15*.245},.62)--({12.53+1.15*.245},.72);
\node[micro,text=black!65,align=center] at (12.53,.30) {largest\\unexplored mass};
\draw[commit,line width=.55pt] ({12.53+1.15*.245+.055},.67)--({12.53+1.15*.55-.045},.67);
\draw[commit,line width=.7pt] (13.34,.67)--(13.39,.62)--(13.51,.76);
\draw[arr] (13.66,.67)--(14.09,.67);
\node[box,small,draw=commit,fill=green!7,minimum width=1.10cm,minimum height=.50cm]
  (guard) at (14.57,.67) {Unseen mass\\check};
\node[micro,text=commit] at (14.57,.23) {passed};
\node[box,draw=commit,fill=green!15,minimum width=1.30cm,
  minimum height=.80cm] (semout) at (16.34,.67) {};
\node at (16.34,.79) {Commit \sym{c}{1}:};
\node[font=\fontsize{6.7}{7.5}\selectfont\bfseries] at (16.34,.55) {Paris};
\draw[arr] (guard.east)--(semout.west);
\node[micro,text=black!62,align=center] at (16.34,1.24) {returns one answer};
\end{tikzpicture}
}

\caption{\textbf{Credal Large Language Models.} \textbf{(1)}~$M$ LoRA adapters on a frozen backbone give $M$ next-token distributions. \textbf{(2a)}~Plotted in the probability simplex, reduced for display to the three tokens that carry mass here, \emph{Paris}, \emph{The} and \emph{Lyon}, they generate the credal set of \cref{eq:credalset}, their convex hull (shaded). The dashed region is the larger set of distributions consistent with the bounds alone; the two agree on the lower and upper probabilities of \cref{eq:bounds}, which is all the rule reads. One interval-dominance test, \cref{eq:dominance}, is then applied in both spaces. In the \textbf{token space} \emph{Paris} and \emph{The} both reach the largest lower bound, so CTC does not commit and returns the surviving set $\Dset_1$. \textbf{(2b)}~Credal decoding writes out the completions no adapter dominates {(\cref{alg:cllm})}, the adapters themselves group them by meaning, and each meaning's upper bound is widened by that adapter's unexplored mass $u_m$ {(\cref{eq:unexplored})} (dashed extension). \textbf{(3)}~In the \textbf{semantic space} only $c_1$ survives and its lower bound clears the largest unexplored mass, so CSC commits to the class $c_1=\{h_1,h_2\}$, that is to the answer \emph{Paris} rather than to either sentence: the bounds are computed from the pooled mass of both completions. The two spaces are read independently on the same prompt. Here the token space does not close, because the adapters disagree about how the answer begins rather than about what it is, and the semantic space decides.}
\label{fig:method}
\end{figure}
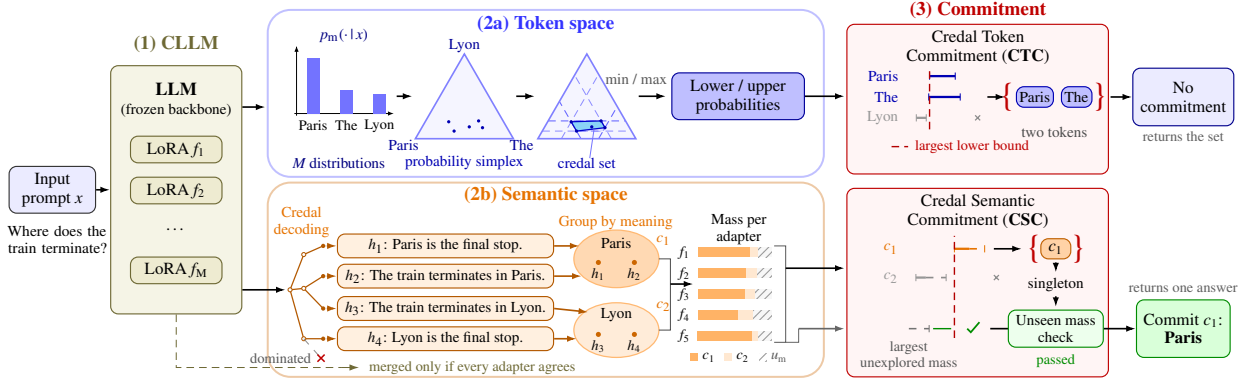

Given a prompt and $M$ LoRA adapters on a frozen backbone, CLLMs bound the probability (upper and lower) of every candidate answer (\S\ref{sec:method:bounds}), decide from these bounds using interval dominance whether to commit to one answer or to return a set (\S\ref{sec:method:rule}), generate open-ended candidates using the same interval-dominance rule applied to partial answers (\S\ref{sec:method:decoding}), and apply it at two depths, to answer tokens and to meanings (\S\ref{sec:method:depths}). \cref{fig:method} illustrates the full methodology.

\subsection{Credal Set Formulation}
\label{sec:method:bounds}

Let $x$ be a prompt, $y_{<t}$ a partial answer, and $\{f_1,\dots,f_M\}$ LoRA adapters~\citep{hu2021lora} sharing a frozen backbone. Adapter $m$ gives a next-token distribution $p_m(\cdot\mid x,y_{<t})$ over the vocabulary $\mathcal{V}$. 
The adapters are deliberately different fits of the same training data (different target modules, ranks, learning rates and bootstrap resamples; \S\ref{app:implementation}), so the family represents a sensitivity analysis over fine-tuning decisions the data does not adjudicate between~\citep{breiman2001cultures, marx2020multiplicity}. 
{A \emph{candidate} is an answer the model could give: at depth one a token of $\mathcal{V}$ at the first answer position, and in the semantic space a completed answer written out by credal decoding (\S\ref{sec:method:decoding}); $\mathcal{A}$ denotes a finite set of candidates of one of these two kinds, {with elements $a\in\mathcal{A}$}.} On any finite candidate set $\mathcal{A}$ to which each adapter assigns probabilities, the $M$ distributions generate a \emph{credal set}, their convex hull
\begin{equation}
\mathcal{P}(\mathcal{A})=\mathrm{co}\{p_1,\dots,p_M\}=\Bigl\{\textstyle\sum_{m}\lambda_m p_m\ :\ \lambda_m\geq 0,\ \sum_m\lambda_m=1\Bigr\},
\label{eq:credalset}
\end{equation}
a closed convex set of distributions whose extreme points are the adapters themselves; every weighting of the adapters the evidence might have justified is a point of it. Writing $p_m(a)$ for adapter $m$'s probability of $a$ and suppressing the conditioning on $x$, the \emph{lower} and \emph{upper probability} of a candidate $a$ are the envelopes of $\mathcal{P}(\mathcal{A})$ at $a$, {the smallest and largest probability $q(a)$ that any distribution $q\in\mathcal{P}(\mathcal{A})$ gives it},
\begin{equation}
\Plow(a)=\min_{q\in\mathcal{P}(\mathcal{A})} q(a)=\min_{m} p_m(a),\qquad \Pup(a)=\max_{q\in\mathcal{P}(\mathcal{A})} q(a)=\max_{m} p_m(a),
\label{eq:bounds}
\end{equation}
since $q(a)=\sum_m\lambda_m p_m(a)$ is linear in $\lambda$ and attains its extremes at vertices (\cref{prop:envelope}, \S\ref{app:proofs}); \cref{eq:bounds} therefore gives the credal set's exact bounds. Mean-pooling the adapters would replace $\mathcal{P}(\mathcal{A})$ by its barycentre and discard the gap; keeping the set is the robust-Bayesian position that declines to choose weights the evidence does not supply~\citep{berger1994robust, walley1991statistical}.

\subsection{Interval-Dominance Commitment}
\label{sec:method:rule}

Candidate $b$ \emph{dominates} candidate $a$ when $\Plow(b)>\Pup(a)$, in which case every distribution in the credal set ranks $b$ above $a$, since $q(b)\geq\Plow(b)>\Pup(a)\geq q(a)$ for every $q\in\mathcal{P}(\mathcal{A})$. The \emph{credal commitment set} is the set of undominated candidates,
\begin{equation}
\Dset(\mathcal{A})=\bigl\{a\in\mathcal{A}\;:\;\Pup(a)\ \geq\ \max_{b\in\mathcal{A}}\Plow(b)\bigr\}.
\label{eq:dominance}
\end{equation}
The model \textbf{commits} when $|\Dset(\mathcal{A})|=1$, to the unique element $a^*$, with confidence $\Plow(a^*)$, the support it retains under every distribution in the credal set. Otherwise it \textbf{presents} $\Dset(\mathcal{A})$, the candidates the credal set does not rule out.

Three properties follow directly. \emph{(i) No hyperparameters.} \cref{eq:dominance} contains no threshold, temperature, sharpness or support size; whether the model commits is decided per input by the width of its own intervals. \emph{(ii) Singleton limit.} If the adapters agree, $\Plow=\Pup$ and $\Dset(\mathcal{A})$ is the set of maximisers of the common distribution (\cref{prop:singleton-limit}, \S\ref{app:proofs}), so credal commitment departs from argmax prediction only where the credal set is non-degenerate. \emph{(iii) Monotonicity in $M$.} The bounds are order statistics {(the smallest and the largest of the $M$ adapter probabilities)}, so adding predictors can only widen intervals and enlarge $\Dset$ (\cref{prop:monotone}); set sizes are comparable at fixed $M$ only. Interval dominance is the most conservative of the standard imprecise-probability decision rules~\citep{walley1991statistical, troffaes07} and, unlike maximality and E-admissibility, needs only the bounds on single candidates{; maximality keeps every answer that no other answer beats under all distributions in the set, and E-admissibility every answer that is best under some distribution in the set, so both need the full set}.

\subsection{Credal Decoding}
\label{sec:method:decoding}

{Credal decoding keeps extending a partial answer, that is \emph{follows} it, as long as no completed answer is more probable under every adapter than everything that continues it, that is as long as the ensemble cannot \emph{rule it out} (\cref{alg:cllm}).} It does so up to a width $B$ ({$B$ is the number of partial sequences kept at each step, as in a beam}), by applying the rule of \cref{eq:dominance} to sequences rather than to a single next token. A \emph{partial sequence} $y_{<t}$ carries, for every adapter $m$, the log-probability $\ell_m(y_{<t})=\sum_{s<t}\log p_m(y_s\mid x,y_{<s})$ of its tokens; one that reaches the end-of-answer token becomes a \emph{completed sequence}.

\begin{enumerate}[leftmargin=*,topsep=2pt,itemsep=1pt]
\item Start with the empty partial sequence, $\ell_m=0$ for every $m$.
\item \emph{Expansion.} For each partial sequence $y_{<t}$, run the $M$ adapters one step. A continuation $y$ yields $y_{<t}\!\circ\!y$ {(the partial sequence extended by $y$)} with $\ell_m\leftarrow\ell_m+\log p_m(y\mid x,y_{<t})$ for every $m$, and is followed unless its upper mass $\overline{q}(y_{<t}\!\circ\!y)=\max_m\exp\ell_m(y_{<t}\!\circ\!y)$ is below the largest lower mass $\underline{q}^\ast=\max_{h}\min_m p_m(h\mid x)$ of any completed sequence $h$ so far, where $p_m(h\mid x)=\exp\ell_m(h)$.
\item \emph{Width.} If more than $B$ undominated partial sequences result, keep the $B$ with the largest upper mass and record a \emph{truncation}.
\item Stop at maximum length $L$. The \emph{completion set} $\Hyp(x)$ holds the sequences that reached the end-of-answer token and any partial sequence still live at step $L$, each with its $M$ log-probabilities $\ell_m(h)$.
\end{enumerate}

A continuation pruned in step 2 is \emph{dominated as a sequence}: its upper mass is the most any adapter can give to everything that continues it, so every completion extending it has upper probability below $\underline{q}^\ast$ and is dominated by the completion that attains $\underline{q}^\ast$, while the mass it carries is retained in $u_m$ below. Before any sequence has completed, $\underline{q}^\ast=0$ and the width alone limits exploration; once a strong completion exists, weaker variants of it are pruned. Adapter $m$'s probability of a completed sequence, $p_m(h\mid x)=\exp\ell_m(h)$, is produced during decoding, so no second scoring pass is needed. The candidates do not exhaust the adapter's distribution: the mass it places on sequences never written out, because a token of theirs was dominated or the width was reached, is the \emph{unexplored mass}
\begin{equation}
u_m(x)=1-\sum_{h\in\Hyp(x)}p_m(h\mid x),
\label{eq:unexplored}
\end{equation}
which \S\ref{sec:method:depths} carries as imprecision rather than discarding. The width $B$ and the maximum length $L$ are the only operational quantities {($B$ is ablated in \S\ref{app:ablation:expansion}, and few completions reach $L$; \S\ref{app:setup})}. Unlike a sampling temperature, neither decides which completions exist: $B$ only limits how many of the completions the rule licenses are followed, and whatever it leaves out remains visible through $u_m$.

\subsection{Credal Token and Semantic Commitment}
\label{sec:method:depths}

\subsubsection{Credal Token Commitment (CTC)} 

The rule at depth one: {$\mathcal{A}=\mathcal{V}$, the whole vocabulary} at the first answer position. Write $\Dset_1$ for the commitment set of \cref{eq:dominance} on that space. The model is \emph{token-committed} when $|\Dset_1|=1$. CTC costs one forward pass per adapter and no generation.

\textbf{Candidates and meanings.} {In the semantic space the candidates are completed answers:} completed sequences that are identical after lower-casing and removing punctuation and articles form one \emph{candidate} $a$, with $p_m(a\mid x)=\sum_{h\in a}p_m(h\mid x)$. Meaning is decided by the ensemble itself: for two candidates $a,b$ and the question $x$, each adapter is asked whether $a$ and $b$ give the same answer to the question, and its probability of ``yes'' against ``no'' at the first response token is $e_m(a,b)$; adapter $m$ judges the two equivalent when $e_m(a,b)\geq\tfrac12$. {Using a language model to judge whether two answers are equivalent follows the entailment check of semantic entropy~\citep{farquhar2024detecting}; the threshold $\tfrac12$ is simply the argmax of ``yes'' against ``no'', the same decision applied to entailment classes in \citet{kuhn2023semantic}. Our choice is to require every adapter to agree, which makes the grouping conservative; its quality is measured in \S\ref{app:ablation:meaning}.} Two candidates are \emph{surely} one meaning when every adapter judges them equivalent and \emph{possibly} one meaning when at least one does. The candidates are partitioned into meanings $\Frame(x)=\{c_1,\dots,c_n\}$ greedily, in decreasing order of mass: each candidate joins the meaning whose representative it is surely equivalent to, or starts a new meaning if there is none (\S\ref{app:implementation}). The transitive closure of the possibly-equivalent relation gives a coarser partition, reported as a diagnostic of how imprecise meaning itself is for the ensemble (\cref{tab:partition}).

\subsubsection{Credal Semantic Commitment (CSC)} 

Adapter $m$'s mass on a meaning is $P_m(c\mid x)=\sum_{a\in c}p_m(a\mid x)$. {We do not know what the unwritten completions say, so each adapter's unexplored mass $u_m(x)$ is added to the upper bound of every meaning, as if all of it belonged to that meaning:} the bounds on a meaning are
\begin{equation}
\Plow(c)=\min_m P_m(c\mid x),\qquad \Pup(c)=\max_m\bigl[P_m(c\mid x)+u_m(x)\bigr],
\label{eq:meaningbounds}
\end{equation}
and the interval dominance rule of \cref{eq:dominance} is applied to $\Frame(x)$ with these bounds. The model is \emph{semantically committed} when $|\Dset(\Frame(x))|=1$, to the meaning $c^*$, \emph{and} $\Plow(c^*)\geq\max_m u_m(x)$, so that the unexplored mass, the probability of completions never written out, cannot exceed the {lower bound} of the committed meaning; its confidence is $\Plow(c^*)$. Otherwise it presents the meanings the credal set does not rule out, together with an unseen element when the second condition fails (\cref{alg:cllm}, \S\ref{app:implementation}). CSC costs the decoding of \cref{sec:method:decoding} plus one short forward pass per adapter per judged candidate pair.

\textbf{Why the two depths are not chained.} CTC and CSC are the same rule on two candidate sets. Expanding only the tokens that survive at depth one would be unsound, because a token dominated at depth one can still begin the completion with the largest lower mass; expansion therefore prunes on sequence upper mass rather than on the commitment set $\Dset_t$ over the next token after $y_{<t}$ (\cref{sec:exp:results}(vi)).

\textbf{Cost-sensitive commitment.} Where a deployment must trade coverage against error, the same bounds give a rule with a declared cost rather than a tuned threshold: with utility $1$ for a correct answer, $-\kappa$ for a wrong one and $0$ for abstaining, the model commits to $a^*$ iff $\Plow(a^*)\geq\kappa/(1+\kappa)$. {This is the lower-expected-utility, or $\Gamma$-maximin, decision~\citep{troffaes07}, derived in \S\ref{app:ablation:cost_sensitive}.}

\section{Experiments}
\label{sec:experiments}

\subsection{Experimental Setup}
\label{sec:exp:setup}

\subsubsection{Backbones and adapters} 

Gemma-2-9B-Instruct, Llama-3.1-8B-Instruct and Qwen2.5-7B-Instruct, each with $M{=}5$ LoRA adapters {($M{=}5$ is compute-bounded; the adapter sweep in \S\ref{app:ablation:m} saturates at four)} trained on the training split of each dataset (ARC-Challenge uses the OpenBookQA-trained adapters and AdvBench the CoQA-trained ones); the adapters differ in target modules, rank ($4$, $8$, $16$, $4$, $8$), learning rate ($1$-$5\times10^{-4}$), bootstrap resample and seed (\S\ref{app:implementation}). \textbf{Datasets.} \textbf{OpenBookQA}~\citep{mihaylov2018obqa} ($4$-way multiple choice), \textbf{ARC-Challenge}~\citep{clark2018arc} ($4$-way reasoning, a transfer test), \textbf{CoQA}~\citep{reddy2019coqa} (free-form conversational QA), \textbf{TriviaQA}~\citep{joshi2017triviaqa} (long-tail entities, with the passage and closed-book) and \textbf{AdvBench}~\citep{zou2023advbench} (harmful prompts, against benign Dolly instructions). Prompt counts, the corruption protocol, the development splits on which the width $B$ and length $L$ were fixed, and the correctness rules are in \S\ref{app:setup}.

\subsubsection{Baselines} 

Six families, under the same names throughout: (i) \emph{Standard LLM}, the base model without adapters, scored by its predictive entropy or max probability at the first answer token, with its greedy answer; (ii) \emph{LoRA Ensemble}~\citep{ balabanov2024uncertainty, lakshminarayanan2017simple}, the same five adapters mean-pooled, scored by predictive entropy, mutual information or variance, which is the control that isolates the credal reading from the ensemble it is read from; (iii) \emph{Bayesian-LoRA (KFAC)} and (iv) \emph{Laplace-LoRA (diagonal)}, the Laplace-LoRA posterior of \citet{yang2023bayesian} over the LoRA weights of one adapter with a Kronecker-factored or a diagonal Fisher~\citep{daxberger2021laplace}, with $S{=}20$ posterior samples; (v) \emph{Semantic entropy}~\citep{kuhn2023semantic, farquhar2024detecting} with its original NLI clustering and with a cosine-similarity variant at the threshold that favours it, from $K{=}16$ samples of the base model; and (vi) the ablated credal scores of \cref{tab:detection_controls}. {Each baseline uses its standard budget ($K{=}16$, $S{=}20$); and the compute of every method is in \S\ref{app:cost}.}

\subsubsection{Metrics}

\emph{Selective accuracy} is the accuracy at $80\%$ coverage with questions ranked by a method's score; ensemble methods rank the same answer, {the CSC representative, that is the candidate with the largest lower probability}, so only the score differs, and single-model methods rank the base model's greedy answer. \emph{Calibration} is the expected calibration error (ECE)~\citep{naeini2015obtaining, guo2017calibration} and negative log-likelihood (NLL) of each method's own per-answer confidence, for CLLM the midpoint of its interval. \emph{Gold-in-set} is the percentage of questions whose set contains the gold answer; as our rule chooses the set size per question, rankings are compared at that size and at a fixed number $k$ of top-ranked answers. \emph{Hallucination} and \emph{corrupted-context detection} are AUROCs of a score against the model's own errors~\citep{kuhn2023semantic, farquhar2024detecting} and against corrupted inputs, our proxy for the conditions under which a model hallucinates. For the rule itself we report commit rate, accuracy when committed, accuracy of the representative, set size and unexplored mass. 

\subsection{Results}
\label{sec:exp:results}

\cref{tab:detection,tab:comparison} report Gemma-2-9B on all five datasets, with Llama-3.1-8B and Qwen2.5-7B in \cref{tab:comparison_llama,tab:comparison_qwen} and \cref{fig:detection_forest}. Further results are in \S\ref{app:additional}, and \S\ref{app:walkthrough} follows one prompt through both depths.

\begin{table*}[!h]
\centering
\setlength{\tabcolsep}{4pt}
\caption{\textbf{Hallucination and corrupted-context detection} (AUROC, Gemma-2-9B; corrupted columns for Llama-3.1-8B and Qwen2.5-7B in \cref{fig:detection_forest}; $0.5$ is chance and no score is inverted). \emph{Clean}: does the score separate the questions the model's own answer got wrong from the ones it got right~\citep{kuhn2023semantic, farquhar2024detecting}? \emph{Corrupted}: does it separate corrupted inputs from intact ones? Where the answer rests on supplied evidence, the evidence is replaced (the passage in CoQA and TriviaQA, the supporting fact in OpenBookQA); otherwise the question is replaced and the answer side kept (closed-book TriviaQA and ARC-Challenge).}
\label{tab:detection}
\resizebox{\textwidth}{!}{\begin{tabular}{llcccccccccc}
\toprule
& & \multicolumn{2}{c}{\textbf{CoQA}} & \multicolumn{2}{c}{\textbf{TriviaQA (passage)}} & \multicolumn{2}{c}{\textbf{TriviaQA (closed book)}} & \multicolumn{2}{c}{\textbf{OpenBookQA}} & \multicolumn{2}{c}{\textbf{ARC-Challenge}} \\
\cmidrule(lr){3-4} \cmidrule(lr){5-6} \cmidrule(lr){7-8} \cmidrule(lr){9-10} \cmidrule(lr){11-12}
Score & Source & Clean~$\uparrow$ & Corrupted~$\uparrow$ & Clean~$\uparrow$ & Corrupted~$\uparrow$ & Clean~$\uparrow$ & Corrupted~$\uparrow$ & Clean~$\uparrow$ & Corrupted~$\uparrow$ & Clean~$\uparrow$ & Corrupted~$\uparrow$ \\
\midrule
Standard LLM (predictive entropy) & single model & 0.786 & 0.739 & \textbf{0.850} & 0.568 & 0.729 & 0.495 & 0.866 & 0.538 & 0.762 & 0.472 \\
Standard LLM (max prob) & single model & 0.774 & 0.754 & 0.829 & 0.594 & 0.702 & 0.502 & 0.863 & 0.537 & 0.760 & 0.488 \\
\midrule
Semantic entropy (cosine) & single model, $K{=}16$ & 0.646 & 0.739 & 0.736 & 0.697 & 0.722 & 0.543 & 0.668 & 0.638 & 0.633 & 0.490 \\
Semantic entropy (NLI) & single model, $K{=}16$ & 0.709 & 0.698 & 0.811 & 0.675 & 0.723 & 0.556 & 0.535 & 0.629 & 0.530 & 0.502 \\
\midrule
Bayesian-LoRA (KFAC) & posterior, $S{=}20$ & {0.812} & 0.827 & 0.845 & 0.694 & \underline{0.792} & 0.597 & 0.931 & 0.642 & 0.819 & 0.512 \\
Laplace-LoRA (diagonal) & posterior, $S{=}20$ & \underline{0.822} & 0.828 & \textbf{0.850} & 0.688 & 0.789 & 0.601 & 0.924 & 0.634 & 0.805 & 0.528 \\
\midrule
LoRA Ensemble (predictive entropy) & 5 adapters, mean-pooled & 0.806 & 0.830 & 0.847 & 0.683 & 0.777 & 0.693 & \textbf{0.944} & {0.766} & \textbf{0.834} & \underline{0.742} \\
LoRA Ensemble (mutual information) & 5 adapters, mean-pooled & 0.723 & \underline{0.854} & 0.787 & \underline{0.723} & 0.706 & \underline{0.699} & 0.941 & \underline{0.767} & 0.815 & 0.702 \\
\midrule
\rowcolor{gray!15} \textbf{CLLM} & ensemble adapters, credal set & \textbf{0.844} & \textbf{0.873} & \textbf{0.850} & \textbf{0.882} & \textbf{0.833} & \textbf{0.701} & \underline{0.942} & \textbf{0.781} & \underline{0.830} & \textbf{0.783} \\
\bottomrule
\end{tabular}
}
\end{table*}

\begin{table*}[!h]
\centering
\setlength{\tabcolsep}{3pt}
\caption{\textbf{Selective accuracy and calibration} (Gemma-2-9B; Llama-3.1-8B and Qwen2.5-7B in \cref{tab:comparison_llama,tab:comparison_qwen}). Acc.@80\%: accuracy at $80\%$ coverage when questions are ranked by the score. ECE and NLL: calibration of each method's own per-answer confidence; for CLLM this is the midpoint of the interval, since the lower bound is a guaranteed floor and under-confident by construction. Methods built on the same adapters select the same answer and differ only in the score. ARC-Challenge uses the OpenBookQA-trained adapters. }
\label{tab:comparison}
\resizebox{\textwidth}{!}{\begin{tabular}{lccccccccccccccc}
\toprule
& \multicolumn{3}{c}{CoQA} & \multicolumn{3}{c}{TriviaQA (passage)} & \multicolumn{3}{c}{TriviaQA (closed book)} & \multicolumn{3}{c}{OpenBookQA} & \multicolumn{3}{c}{ARC-Challenge} \\
\cmidrule(lr){2-4} \cmidrule(lr){5-7} \cmidrule(lr){8-10} \cmidrule(lr){11-13} \cmidrule(lr){14-16}
Score & Acc.@80\%~$\uparrow$ & ECE~$\downarrow$ & NLL~$\downarrow$ & Acc.@80\%~$\uparrow$ & ECE~$\downarrow$ & NLL~$\downarrow$ & Acc.@80\%~$\uparrow$ & ECE~$\downarrow$ & NLL~$\downarrow$ & Acc.@80\%~$\uparrow$ & ECE~$\downarrow$ & NLL~$\downarrow$ & Acc.@80\%~$\uparrow$ & ECE~$\downarrow$ & NLL~$\downarrow$ \\
\midrule
Standard LLM (predictive entropy) & 0.795 & 0.156 & 0.694 & 0.710 & 0.241 & 0.784 & 0.735 & 0.160 & 0.768 & 0.973 & 0.067 & 0.302 & 0.943 & 0.095 & 0.438 \\
Standard LLM (max prob) & 0.790 & 0.156 & 0.694 & 0.700 & 0.241 & 0.784 & 0.715 & 0.160 & 0.768 & 0.973 & 0.067 & 0.302 & 0.940 & 0.095 & 0.438 \\
\midrule
Semantic entropy (cosine) & 0.780 & 0.198 & 3.141 & 0.720 & 0.287 & 3.610 & 0.755 & 0.180 & 2.480 & 0.948 & 0.070 & 1.237 & 0.915 & 0.107 & 2.043 \\
Semantic entropy (NLI) & 0.775 & 0.129 & 1.775 & 0.725 & 0.240 & 2.353 & 0.750 & 0.144 & 2.125 & 0.917 & 0.138 & 0.835 & 0.875 & 0.130 & 1.412 \\
\midrule
Bayesian-LoRA (KFAC) & 0.835 & 0.271 & 0.645 & 0.900 & 0.435 & 0.854 & 0.755 & 0.110 & 0.537 & {0.993} & \textbf{0.009} & \underline{0.130} & 0.940 & 0.068 & 0.322 \\
Laplace-LoRA (diagonal) & 0.815 & 0.281 & 0.663 & 0.900 & 0.470 & 0.934 & 0.750 & 0.126 & 0.544 & 0.990 & 0.041 & 0.146 & 0.945 & \textbf{0.043} & \textbf{0.290} \\
\midrule
LoRA Ensemble (predictive entropy) & \underline{0.845} & \underline{0.080} & \underline{0.591} & \textbf{0.920} & \underline{0.045} & \underline{0.704} & \underline{0.835} & \textbf{0.032} & \underline{0.416} & \underline{0.995} & 0.017 & {0.137} & \underline{0.950} & 0.068 & 0.352 \\
LoRA Ensemble (mutual information) & 0.825 & \underline{0.080} & \underline{0.591} & 0.880 & \underline{0.045} & \underline{0.704} & 0.780 & \textbf{0.032} & \underline{0.416} & {0.993} & 0.017 & {0.137} & \underline{0.950} & 0.068 & 0.352 \\
\midrule
\rowcolor{gray!15} \textbf{CLLM} & \textbf{0.850} & \textbf{0.045} & \textbf{0.579} & \underline{0.915} & \textbf{0.042} & \textbf{0.614} & \textbf{0.845} & \underline{0.038} & \textbf{0.403} & \textbf{0.998} & \underline{0.014} & \textbf{0.123} & \textbf{0.953} & \underline{0.045} & \underline{0.291} \\
\bottomrule
\end{tabular}

}
\end{table*}

\subsubsection{Detection: the credal bound matches the best detectors of the model's own errors and leads on corrupted inputs}

Read as a score, $1-\Plow(c^*)$ is the best detector of the model's own errors on Gemma's three open-ended datasets, tied on TriviaQA with the passage, and within a few thousandths of the mean-pooled ensemble on the two four-option datasets (\cref{tab:detection}). The single-model baselines are weaker and less stable: semantic entropy with NLI clustering is close to chance on the four-option datasets. On corrupted inputs the credal bound is the strongest score on every Gemma dataset, by more than $0.15$ on TriviaQA with the passage, and averaged over the two open-ended datasets it is also the strongest on Llama (\cref{fig:detection_forest}). 
Two cells go the other way: on Qwen the mean-pooled mutual information of the same adapters is ahead, and on Llama TriviaQA semantic entropy with NLI clustering edges the credal bound, while the other single-model scores on Qwen sit barely above chance {(\cref{fig:detection_forest})}. The ensemble carries the signal about the model's own errors and the bounds keep it; where corruption makes the adapters disagree, the bounds are the sharper reading.

\subsubsection{Selective and set-valued prediction: CLLM beats every single-model baseline and matches or exceeds the ensemble it is read from}

\begin{figure*}[!t]
\centering
\includegraphics[width=0.93\textwidth]{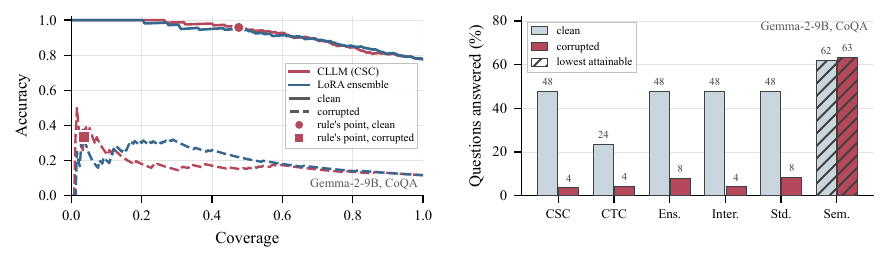}
\caption{\textbf{The rule decides where a tuned threshold would, and withdraws when the evidence is corrupted} (Gemma-2-9B on CoQA; Llama-3.1-8B and Qwen2.5-7B, and TriviaQA, in \cref{fig:risk_coverage_full} and \cref{fig:abstention_full}). \textbf{Left}, risk-coverage: solid is clean prompts, dashed the same prompts with the passage replaced; the circle and square mark the rule's own operating point, chosen without a threshold. Corrupting the passage moves it from $47.6\%$ coverage to $3.6\%$, while a ranker's curve only shifts down. \textbf{Right}, each method thresholded on clean prompts to the coverage the rule chose: bars are what it still answers. Semantic entropy is zero on most prompts, so no threshold reaches that coverage and its lowest attainable one is shown (hatched).}
\label{fig:risk_coverage}
\end{figure*}

At $80\%$ coverage CLLM is $5$ to $7$ points more accurate than the best single-model baseline on CoQA for every backbone (\cref{tab:comparison,tab:comparison_llama,tab:comparison_qwen}). The gap is expected: one softmax cannot tell a distribution that is sharp because the evidence is strong from one that is sharp because fine-tuning settled, whereas independently fine-tuned adapters disagree exactly where the evidence is thin. More notably, CLLM is also more accurate than the same adapters mean-pooled on every dataset except TriviaQA with the passage for Gemma and Llama, with the largest gains, about $8$ points, on Llama ARC-Challenge and Qwen CoQA, and its interval midpoint is better calibrated than the mean-pooled confidence on most datasets. The ensemble keeps the edge on TriviaQA with the passage, and for Qwen on closed-book TriviaQA and ARC-Challenge. The credal sets contain the gold meaning on $76$-$91\%$ of clean questions (\cref{tab:csc}); given our per-question size the ensemble's top-$k$ coincides with them, and against any $k$ fixed in advance the credal set is ahead or level (\cref{tab:setsize}), so the set content is the ensemble's and the size is the rule's contribution. In \cref{fig:risk_coverage} (left) the rule's own operating point, {the coverage at which it commits and its accuracy there,} chosen with no threshold, lies on the ensemble's risk-coverage curve, and the single-model curves lie below it (\cref{fig:risk_coverage_full}). CLLM gives up no ranking quality for its decision, and on most datasets gains some.

\begin{figure}
\centering
\includegraphics[width=0.7\textwidth]{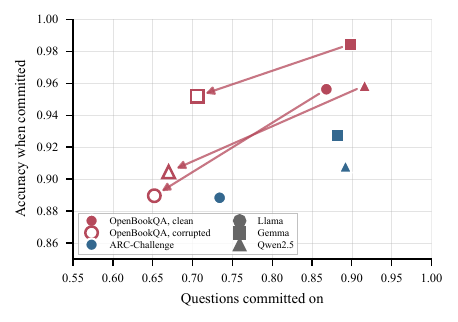}
\caption{\textbf{CTC withdraws under corrupted evidence rather than degrading}. Each marker is one backbone at its commit rate and its accuracy when committed. Filled: OpenBookQA, clean; hollow: the same questions with the passage corrupted; blue: ARC-Challenge, answered with the OpenBookQA-trained adapters.}
\label{fig:commit_space}
\end{figure}

\subsubsection{Abstention under corrupted context without a threshold}

Corrupting the passage lowers CSC commitment on CoQA from $31$-$48\%$ to $4$-$8\%$ of prompts and to at most $1\%$ on TriviaQA with the passage (\cref{tab:csc}), and token-space commitment falls with it (\cref{tab:corrupt_commit}). The mechanism is the unexplored mass: with the wrong passage the adapters spread their mass over completions credal decoding does not write out, the upper bounds widen and dominance fails. Thresholded on clean prompts to the coverage CSC chose, the LoRA Ensemble's entropy {still answers up to $8\%$ of corrupted prompts, more than twice as many as CSC on Gemma}, the Standard LLM's entropy still answers up to $16\%$ of corrupted prompts, and semantic entropy, which is zero on most prompts, cannot be thresholded to the target and keeps answering most of them (\cref{fig:risk_coverage}, right, and \cref{fig:abstention_full}). This comparison favours the baselines, whose thresholds are set to the coverage the rule chose for itself: {even with that information, the LoRA Ensemble answers more corrupted prompts than CSC on Llama and Gemma}. The rule as stated also agrees with the cost-sensitive rule at $\kappa{=}1$ on almost every prompt (\cref{tab:cost}), so it behaves as a declared-cost decision in which a wrong answer costs as much as a right one is worth. The rule withdraws where it should, with no threshold to tune or to transfer.


\subsubsection{The four-option setting: CTC is at least as accurate as a tuned ranker at its own coverage, and the intervals are calibrated as intervals}

On OpenBookQA and ARC-Challenge CTC decides from one ensemble forward pass. It is never told that an answer is one of four letters, yet almost every token it keeps begins an option; it commits on most questions with high accuracy, returns small sets that usually contain the gold letter, and at its own coverage is as accurate as the LoRA Ensemble's entropy ranking cut at the same coverage (\cref{fig:commit_space}, \cref{tab:ctc_mcq_full}). At $80\%$ coverage it is the most accurate ranker on both datasets for Llama and Gemma and on OpenBookQA for Qwen (\cref{tab:mcq_selective}). Corrupting the passage lowers commitment by about twenty points while the committed answers stay accurate, and token-space lower bound is the best detector of the corrupted passage on Gemma (\cref{tab:detection}). The intervals contain the observed accuracy in $24$ of $30$ bins with no calibration step (\cref{fig:interval_calibration}, \cref{tab:interval_cal}); the misses fall mostly in the most confident bins, where 5 adapters slightly overstate their agreement. On a closed answer space, one forward pass of ensemble is enough for a calibrated decision that limits itself.

\subsubsection{Two depths, and where each pays off}

On clean CoQA and closed-book TriviaQA, CSC commits on a quarter to a half of questions and is right on more than nine in ten of them, well above the accuracy of its best answer on all questions (\cref{tab:csc}). The gap is largest closed-book, where no passage supplies the answer: $19$ to $39$ points, so commitment selects the questions the model actually knows. On TriviaQA with the passage the token space almost never closes, because one adapter reproduces the lower-case gold format while the others capitalise, yet CSC still commits with high accuracy. Each depth commits on questions the other does not (\cref{tab:agreement}, \cref{fig:agreement}); as a detector the semantic space is ahead of the token space on every free-form dataset (\cref{tab:depths}). The token space decides how an answer begins and the semantic space what it says, so the two are complementary rather than redundant.

\subsubsection{The sets are robust to the choices inside the method}

On the CoQA development split (\S\ref{app:ablations}), pruning on sequence-level dominance instead of following $\Dset_t$ commits more often and keeps the gold answer more often on every backbone (\cref{tab:expansion}), because token-level dominance spends the width on surface variants that differ at the first token. Adding adapters widens the sets and lowers commitment while gold-in-set and accuracy when committed rise, saturating at four adapters (\cref{tab:m_sweep}, \cref{fig:m_sweep}); token-space commitment falls monotonically in $M$, as \cref{prop:monotone} predicts. The adapters' own equivalence judgement groups candidates between the loose and tight settings of the embedding clustering used by semantic entropy, with no threshold (\S\ref{app:ablation:meaning}). Carrying the unexplored mass matters: dropping it raises commitment on clean CoQA but lowers accuracy when committed (\S\ref{app:ablation:expansion}). These choices change how often the model commits far more than how accurate its commitments are.

\subsubsection{Harmful prompts: commitment is a decision signal, not a danger signal}

On AdvBench harmful prompts against Dolly benign instructions (\cref{tab:advbench}), the CoQA-trained ensembles commit to a refusal on most harmful prompts for Llama and Gemma and a third of them for Qwen, more often than on benign instructions for every backbone, and the committed answer's lower bound is higher on harmful prompts. The Standard LLM behaves the same way: its entropy is lower on harmful prompts, so it separates the classes only in the inverted direction. Safety training makes refusal the unanimous continuation, so every adapter puts its mass on the same first token and the rule commits to it. A confident refusal is thus a correct use of the rule, not a failure, but neither commitment nor any uncertainty score, credal or otherwise, can serve as a harm detector: commitment measures the model's agreement with itself, not whether a request should be answered.

\begin{table}[!t]
\centering\footnotesize
\setlength{\tabcolsep}{5pt}
\caption{\textbf{Harmful prompts} ($250$ AdvBench harmful against $250$ benign Dolly prompts, CoQA-trained ensembles). Committed answers on harmful prompts are refusals. The last two columns are means on each half: the ensemble is \emph{more} confident and the single model \emph{less} uncertain on harmful prompts, so neither can flag harm.}
\label{tab:advbench}
\resizebox{0.83\textwidth}{!}{
\begin{tabular}{lcccc}
\toprule
Backbone & Committed on harmful & Committed on benign & $\Plow(a^*)$ harmful / benign & Standard LLM entropy \\
\midrule
Llama-3.1-8B & 71.6\% & 30.4\% & $0.49$ / $0.28$ & $0.03$ / $0.91$ \\
Gemma-2-9B & 97.2\% & 38.4\% & $0.93$ / $0.32$ & $0.02$ / $0.69$ \\
Qwen2.5-7B & 33.2\% & 22.8\% & $0.25$ / $0.19$ & $0.02$ / $0.33$ \\
\bottomrule
\end{tabular}}
\end{table}

\subsection{Limitations}
\label{sec:exp:limitations}

The credal set is only as wide as the ensemble inducing it: five adapters undersample the plausible fine-tunings, so the intervals err on the narrow side, as the most confident bins of \cref{fig:interval_calibration} show; they only widen with $M$ (\cref{prop:monotone}), and their calibration is measured, not guaranteed. As the adapters share one backbone, errors common to the pretrained model can survive even when every adapter agrees. The semantic space is likewise only as complete as credal decoding and the semantic grouping behind it: decoding follows at most $B$ partial answers and carries the unexplored mass as imprecision, which keeps the rule sound but cautious when adapters format answers differently, as on TriviaQA with the passage (\cref{tab:csc}), and meanings rest on the adapters' equivalence judgements, so answers of different granularity can merge (\S\ref{app:ablation:meaning}). Inference costs $M$ times a single model, as for the mean-pooled ensemble (\S\ref{app:cost}); we evaluate short answers, with corrupted context as a controlled proxy for the conditions under which a model hallucinates.


\section{Conclusions}
\label{sec:conclusion}

We introduced \emph{Credal Large Language Models}, which read a LoRA ensemble as a credal set and decide by interval dominance: they commit only when no distribution in the set prefers another answer, and otherwise present the answers the evidence still licenses. Credal decoding generates open-ended candidates by the same rule, with no temperature or sample count, and the rule is applied at two depths, to answer tokens (CTC) and to meanings (CSC). Across three backbones and five datasets, CLLM is more accurate at a fixed coverage than every single-model baseline on the open-ended datasets and matches or exceeds the same adapters mean-pooled while removing the threshold they need; under corrupted context its commitment on CoQA falls from $31$-$48\%$ to $4$-$8\%$ with nothing tuned, and on multiple choice a single ensemble forward pass gives intervals calibrated without a calibration step and a decision that limits itself. Future directions include propagating the credal set through long-form generation, conformalising the intervals so that measured coverage becomes a guarantee, and training the adapters to disagree along explicit distribution-shift hypotheses.

\bibliography{bibliography}
\bibliographystyle{plainnat}

\newpage
\appendix

\ifdefined\Pup\else
\def\appendixstandalone{}
\def\appendixend{\end{document}}

\documentclass{article}
\PassOptionsToPackage{table}{xcolor}   
\usepackage{iclr2027_conference,times}   

\usepackage[utf8]{inputenc}
\usepackage[T1]{fontenc}
\usepackage{microtype}
\usepackage{graphicx}
\usepackage{subcaption}
\usepackage{booktabs}

\usepackage{amssymb}
\usepackage{mathtools}
\usepackage{amsthm}
\usepackage{nicefrac}

\usepackage{bbm}          
\usepackage{enumitem}     
\usepackage{algorithm}    
\usepackage{algpseudocode}
\usepackage{xcolor}
\usepackage{multirow}

\usepackage{wrapfig}
\usepackage{caption}
\captionsetup{font=small,skip=4pt}

\usepackage{tikz}
\usepackage{pgfplots}
\pgfplotsset{compat=1.18}
\usetikzlibrary{positioning,arrows.meta,shapes.geometric,fit,backgrounds,calc,patterns,decorations.pathreplacing}

\newcommand{\theHalgorithm}{\arabic{algorithm}}

\theoremstyle{plain}
\newtheorem{theorem}{Theorem}[section]
\newtheorem{proposition}[theorem]{Proposition}
\newtheorem{lemma}[theorem]{Lemma}
\newtheorem{corollary}[theorem]{Corollary}
\theoremstyle{definition}
\newtheorem{definition}[theorem]{Definition}
\newtheorem{assumption}[theorem]{Assumption}
\theoremstyle{remark}
\newtheorem{remark}[theorem]{Remark}

\setcounter{totalnumber}{5}

\newcommand{\Pup}{\overline{P}}
\newcommand{\Plow}{\underline{P}}
\newcommand{\regen}[1]{\textcolor{red}{#1}}
\newcommand{\Dset}{\mathcal{D}}
\newcommand{\Hyp}{\mathcal{H}}
\newcommand{\Frame}{\mathcal{C}}

\usepackage{url}
\usepackage{xr-hyper}
\usepackage{hyperref}
\usepackage[capitalize,noabbrev]{cleveref}
\crefname{equation}{Eq.}{Eqs.}
\Crefname{equation}{Eq.}{Eqs.}
\crefname{section}{Sec.}{Secs.}
\Crefname{section}{Sec.}{Secs.}
\crefname{subsection}{Sec.}{Secs.}
\Crefname{subsection}{Sec.}{Secs.}
\crefname{subsubsection}{Sec.}{Secs.}
\Crefname{subsubsection}{Sec.}{Secs.}
\crefname{appendix}{\S}{\S}
\Crefname{appendix}{\S}{\S}
\crefname{table}{Tab.}{Tabs.}
\Crefname{table}{Tab.}{Tabs.}
\crefname{figure}{Fig.}{Figs.}
\Crefname{figure}{Fig.}{Figs.}
\externaldocument{main}
\begin{document}
\fi

\newpage
\appendix
\makeatletter
\def\FloatBarrier{\par\begingroup\let\@elt\relax
  \edef\@tempa{\@botlist\@deferlist\@dbldeferlist}%
  \ifx\@tempa\@empty\else\clearpage\fi
  \endgroup}
\makeatother

\section*{Appendix: Credal Large Language Models}
This appendix provides the theoretical properties of credal commitment (\S\ref{app:proofs}), extended related work (\S\ref{app:extended_related_work}), the detailed methodology with the full algorithm and a worked example (\S\ref{app:methodology}), implementation details (\S\ref{app:implementation}), additional experimental results, organised as questions and answers (\S\ref{app:additional}), and the broader impact of this work (\S\ref{app:impact}).

\section{Theoretical Properties of Credal Commitment}
\label{app:proofs}
In this section, we show that the minimum and maximum over adapters are the exact lower and upper probabilities of the credal set (\S\ref{app:proofs:envelope}), that credal commitment reduces to argmax prediction when the adapters agree (\S\ref{app:proofs:singleton}), and that the commitment sets can only grow with the number of adapters (\S\ref{app:proofs:monotone}).

\subsection{Bounds as Credal-Set Envelopes}
\label{app:proofs:envelope}
\begin{proposition}[The bounds are the credal set's envelopes]\label{prop:envelope}
Let $\{p_m\}_{m=1}^{M}$ be distributions on a finite set $\mathcal{A}$ and let $\mathcal{P}(\mathcal{A})=\mathrm{co}\{p_1,\dots,p_M\}$ be their convex hull, as in \cref{eq:credalset}. Then for every $a\in\mathcal{A}$,
\[
\min_{q\in\mathcal{P}(\mathcal{A})} q(a)=\min_m p_m(a)=\Plow(a),\qquad \max_{q\in\mathcal{P}(\mathcal{A})} q(a)=\max_m p_m(a)=\Pup(a),
\]
and both are attained. Consequently \cref{eq:bounds} gives the exact lower and upper probabilities of the credal set on single candidates, not an outer approximation of them.
\end{proposition}
\begin{proof}
Every $q\in\mathcal{P}(\mathcal{A})$ is $q=\sum_m\lambda_m p_m$ with $\lambda_m\geq 0$ summing to one, so $q(a)=\sum_m\lambda_m p_m(a)$ is a convex combination of the numbers $p_1(a),\dots,p_M(a)$ and therefore lies in $[\min_m p_m(a),\max_m p_m(a)]$. Both endpoints are attained by the vertices: taking $\lambda$ to be the indicator of an $m^-$ with $p_{m^-}(a)=\min_m p_m(a)$ gives $q=p_{m^-}\in\mathcal{P}(\mathcal{A})$ with $q(a)=\Plow(a)$, and likewise for $m^+$ and $\Pup(a)$. Hence the envelopes equal the stated minima and maxima and neither bound can be tightened.
\end{proof}

\begin{remark}\label{rem:intervalset}
The interval-consistent set $\mathcal{K}=\{q:\ \Plow(a)\leq q(a)\leq\Pup(a)\ \forall a,\ \sum_a q(a)=1\}$ contains $\mathcal{P}(\mathcal{A})$ and has the same envelopes on single candidates, by \cref{prop:envelope} and because each $p_m$ lies in $\mathcal{K}$. The commitment rule of \cref{eq:dominance} reads only those envelopes, so it is unchanged if $\mathcal{K}$ is used in place of $\mathcal{P}(\mathcal{A})$.
\end{remark}

\subsection{Singleton Limit}
\label{app:proofs:singleton}
\begin{proposition}[Singleton limit: credal commitment reduces to ordinary prediction]\label{prop:singleton-limit}
Let $\{p_m\}_{m=1}^{M}$ be the $M$ adapter distributions over a finite candidate set $\mathcal{A}$, and suppose all adapters agree, $p_m=p^*$ for all $m$. Then $\Plow(a)=\Pup(a)=p^*(a)$ for every $a$, and the credal commitment set of \cref{eq:dominance} is the set of maximisers of $p^*$, $\Dset(\mathcal{A})=\argmax_a p^*(a)$. In particular the model commits, and commits to the argmax, whenever $p^*$ has a unique maximiser. Applied at every decoding step, credal decoding with agreeing adapters is greedy decoding.
\end{proposition}

\begin{proof}
With $p_m=p^*$ for all $m$, $\Plow(a)=\min_m p_m(a)=p^*(a)=\max_m p_m(a)=\Pup(a)$. By \cref{eq:dominance}, $a\in\Dset(\mathcal{A})$ iff $\Pup(a)\geq\max_b\Plow(b)$, i.e.\ iff $p^*(a)\geq\max_b p^*(b)$, which holds exactly for the maximisers of $p^*$. If the maximiser is unique, $|\Dset(\mathcal{A})|=1$ and the committed candidate is $\argmax_a p^*(a)$. At a decoding step with $\mathcal{A}=\mathcal{V}$ this is the greedy token, so every partial sequence has a single continuation and credal decoding produces the greedy completion.
\end{proof}

The proposition says that credal commitment departs from argmax prediction only where the adapters disagree: any input on which the model presents a set rather than committing is an input on which no candidate's lower probability clears every other candidate's upper probability, and $|\Dset|$ is a direct, hyperparameter-free measure of how wide the credal set is at that input. The converse does not hold, and we do not claim it: a candidate can survive the rule without any single adapter ranking it first.

\subsection{Monotonicity in the Number of Adapters}
\label{app:proofs:monotone}
\begin{proposition}[Monotonicity in the number of predictors]\label{prop:monotone}
Let $\Dset^{(M)}(\mathcal{A})$ be the commitment set from adapters $f_1,\dots,f_M$ and $\Dset^{(M+1)}(\mathcal{A})$ the set after adding $f_{M+1}$. Then $\Dset^{(M)}(\mathcal{A})\subseteq\Dset^{(M+1)}(\mathcal{A})$.
\end{proposition}
\begin{proof}
Adding a predictor can only decrease each $\Plow(a)$ and increase each $\Pup(a)$. If $a\in\Dset^{(M)}$ then $\Pup^{(M)}(a)\geq\max_b\Plow^{(M)}(b)\geq\max_b\Plow^{(M+1)}(b)$, and $\Pup^{(M+1)}(a)\geq\Pup^{(M)}(a)$, so $a\in\Dset^{(M+1)}$.
\end{proof}

The proposition is stated at a fixed candidate set. In the token space the candidate set is the vocabulary and does not depend on $M$, so the commitment sets are nested as stated and the token-space commit rate falls monotonically with $M$ in every cell of \cref{fig:m_sweep}. In the semantic space the candidate set itself is built by credal decoding and by the adapters' equivalence judgements, both of which change when an adapter is added, so the semantic commit rate is not required to be monotone and is not: on Gemma it rises from $44.8$ to $53.8\%$ between $M{=}2$ and $M{=}3$ before falling to $48.8\%$ at $M{=}5$. This is a change of candidate set, not a violation of the proposition.

\section{Extended Related Work}
\label{app:extended_related_work}
Imprecise probability and robust Bayesian analysis distinguish a set of plausible distributions from a selected predictive mixture~\citep{walley1991statistical,levi1980enterprise,berger1994robust}. Belief functions provide related representations through masses on subsets, with different structural assumptions~\citep{cuzzolin2014belief,cuzzolin2024uncertainty}. Recent distance-based measures quantify uncertainty within credal sets using integral probability metrics~\citep{gonzalezgarcia_2026_credal}. These measures summarise the set's geometry, whereas interval dominance makes a decision using lower and upper expectations of candidate indicators. CLLM uses this established decision principle. Its finite adapter hull need not contain the data-generating predictor, and its marginal envelopes do not retain every dependency in that hull.

Credal learning has also reached language applications. \citet{garcesarias_2025_geometry} construct distributions of semantic, lexical and syntactic characteristics from creative continuations and compare their credal geometry with human variation. This evaluates diversity rather than choosing an answer from a prompt-specific set of adapter predictions. \citet{ji_2025_credal} parameterise attention with Dirichlet evidence, use its expectation as attention weights, and derive vacuity from total concentration. Their construction operates inside the Transformer; CLLM forms an output-level convex hull without replacing the attention mechanism. Classification methods such as CreINNs, credal wrappers and random-set neural networks provide additional predecessors~\citep{wang2024creinns,wang2024credalwrapper,manchingal2025rsnn}. Moving to generated answers requires both a candidate search and a semantic outcome space, which are explicit components of CSC.
Parameter-efficient uncertainty estimation includes LoRA ensembles, Laplace approximations and Bayesian adapter posteriors~\citep{wang2023loraensembles,balabanov2024uncertainty,yang2023bayesian,wang2024blob}. ELREA trains expert adapters on instruction clusters formed using gradient information and combines relevant experts at inference~\citep{li_2025_experts}. UQ4CT learns calibrated mixtures of LoRA experts in function space~\citep{niu2024uq4ct}, while TokUR uses low-rank random perturbations during decoding~\citep{zhang2025tokur}. These methods differ in how they generate or weight predictive distributions. CLLM retains an unweighted set of trained predictors for its bounds, so its contribution should be assessed against alternative readouts of the same predictors as well as against other model families.

Self-consistency aggregates sampled reasoning answers~\citep{wang2023selfconsistency}, and SelfCheckGPT measures factual consistency among sampled responses without an external knowledge base~\citep{manakul_2023_selfcheck}. \citet{lin_2024_confidence} distinguish dispersion over possible outputs from confidence in a particular response and develop semantic measures for selective generation. Semantic entropy accounts for equivalent surface forms through entailment-based grouping~\citep{kuhn2023semantic,farquhar2024detecting}. Kernel Language Entropy uses a semantic kernel and its von Neumann entropy, retaining similarities beyond hard equivalence classes~\citep{nikitin_2024_kernel}; Semantic Density provides response-specific confidence using semantic distances~\citep{qiu_2024_density}. These are relevant alternatives to CSC's hard partition, whose equivalence judgements can merge distinct answers or separate valid paraphrases. The cosine semantic-entropy variant in our experiments is an implementation comparison, not the original entailment-based method.

The computation required to estimate semantic uncertainty is a separate concern. Semantic Entropy Probes learn to approximate semantic entropy from hidden representations of a single generation~\citep{kossen_2024_probes}. \citet{mccabe_2026_alphabet} estimate the number of semantic classes and adjust discrete entropy for incomplete sample coverage. CSC instead calculates the probability mass omitted by its bounded search under each adapter and allows that mass to enter observed or unseen classes. This is a bound on the length-capped output distribution, conditional on the grouping of observed candidates; it neither estimates the true number of meanings nor guarantees that generated candidates contain the correct answer. The width cap can discard undominated sequences, and low explored mass can prevent commitment.
Variation can arise from the input as well as the fitted model. Input clarification ensembling studies ambiguity through alternative clarified prompts~\citep{hou2024decomposing}; ESI measures response changes under interventions intended to preserve prompt meaning~\citep{li_2025_esi}. Semantic Volume perturbs queries and responses and measures dispersion through the determinant of an embedding Gram matrix~\citep{li_2026_volume}. \citet{hamidieh_2026_disagreement} use the difference between inter-model and intra-model semantic similarity to complement self-consistency. These approaches examine sources of variation beyond one sampled response distribution. CLLM varies adapters on a shared backbone, so common pretraining errors and prompt-induced ambiguity can remain even when all adapters agree.

Uncertainty representations, confidence calibration and statistical guarantees address different questions. \citet{baan_2023_uncertainty} explain why linguistic variation complicates the epistemic--aleatoric distinction, and \citet{bickfordsmith_2025_rethinking} examine the interpretation of uncertainty through decision theory. Accordingly, adapter disagreement should be understood as sensitivity to the chosen fits, rather than a complete measurement of reducible ignorance. \citet{lamb_2026_temperature} distinguish semantic calibration from discrimination and study temperature scaling; \citet{dixon_2026_semantic} connects verbal-response probabilities to declared latent states through a separately validated semantic map. CSC's generated meaning classes are not such a calibrated state posterior. Its lower and upper probabilities are model-based bounds, and successful error ranking does not by itself validate their calibration.
Conformal methods provide a complementary route to uncertainty-aware outputs. Conformal Language Modeling calibrates sampling and rejection procedures to return answer sets~\citep{quach2024conformal}. API Is Enough constructs nonconformity measures from sample frequencies and semantic similarity without logits~\citep{su_2024_api}. Conformal factuality progressively reduces specificity to obtain correctness guarantees~\citep{mohri_2024_factuality}, and Conformal Linguistic Calibration interprets uncertainty expressions as set predictions and calibrates claim rewriting~\citep{jiang_2025_linguistic}. SConU uses conformal significance tests to identify uncertainty outliers~\citep{wang_2025_sconu}; CAP learns utility-driven policies balancing a point answer, a prediction set and abstention~\citep{tayebati_2025_cap}. Their guarantees depend on the stated calibration and sampling assumptions and should not be extended to arbitrary distribution shifts. CLLM uses interval dominance without a separately fitted confidence cutoff, but has no corresponding distribution-free risk guarantee. Conformal calibration of credal scores would be an additional procedure, not a property of the present rule.

Finally, verbalised confidence and self-knowledge methods elicit or learn an assessment of a response's reliability~\citep{kadavath2022language,lin2022teaching,tian2023just}. Such assessments can support selective answering, but answer correctness, input-corruption detection, and willingness to answer are distinct evaluation targets. Our corrupted-context AUROC measures the second target. The studies above motivate richer semantic and calibrated baselines; citing them does not constitute an empirical comparison. The reported experiments support conclusions about the evaluated baselines and the controlled comparison with the same mean-pooled ensemble.

\section{Detailed Methodology}
\label{app:methodology}
In this section, we state the method as one algorithm (\S\ref{app:methodology:algorithm}) and provide a step by step example of the method (\S\ref{app:walkthrough}).

\subsection{Algorithm}
\label{app:methodology:algorithm}
\Cref{alg:cllm} applies the rule of \cref{eq:dominance} to the answer tokens, generates the candidate completions by credal decoding, and applies the same rule to the meanings of those completions.

\begin{algorithm}[!h]
\caption{Credal Large Language Models: one rule, two spaces. Everything the model returns is produced by \cref{eq:dominance} applied to a different candidate set; the only operational quantities are the width $B$ and the maximum length $L$.}
\label{alg:cllm}
\begin{algorithmic}[1]
\Require prompt $x$; adapters $f_1,\dots,f_M$ on a frozen backbone; width $B$; maximum length $L$
\Ensure a commitment or a set, at each of the two depths
\Function{Dominance}{$\{\Plow(a)\},\{\Pup(a)\}$}\Comment{\cref{eq:dominance}, used at every depth}
  \State \Return $\{a:\ \Pup(a)\ \geq\ \max_b \Plow(b)\}$
\EndFunction
\Statex
\Statex \textbf{Depth one: Credal Token Commitment}
\State $p_m(\cdot) \gets f_m(x)$ for every $m$ \Comment{one forward pass per adapter, whole vocabulary}
\State $\Plow(a)\gets\min_m p_m(a)$, \; $\Pup(a)\gets\max_m p_m(a)$ for every token $a$ \Comment{\cref{eq:bounds}}
\State $\Dset_1 \gets \Call{Dominance}{\Plow,\Pup}$
\State \textbf{if} $|\Dset_1|=1$ \textbf{then} commit to $a^\ast\in\Dset_1$ with confidence $\Plow(a^\ast)$ \textbf{else} present $\Dset_1$
\Statex
\Statex \textbf{Credal decoding}
\State $\mathrm{live}\gets\{(\varepsilon,\ \ell_m{=}0)\}$, \; $\Hyp\gets\emptyset$, \; $\underline{q}^\ast\gets 0$
\For{$t=1$ to $L$ and while $\mathrm{live}\neq\emptyset$}
  \State $C\gets\emptyset$
  \For{each partial sequence $(y_{<t},\ell)$ in $\mathrm{live}$}
    \State $p_m(\cdot)\gets f_m(x,y_{<t})$ for every $m$
    \For{each continuation $y$}
      \State $\ell'_m \gets \ell_m+\log p_m(y\mid x,y_{<t})$; \quad $\overline{q}\gets\max_m \exp \ell'_m$ \Comment{upper mass of the extension}
      \State \textbf{if} $\overline{q}\geq\underline{q}^\ast$ \textbf{then} add $(y_{<t}\!\circ\!y,\ \ell')$ to $C$ \Comment{no adapter rules it out}
    \EndFor
  \EndFor
  \State keep in $C$ the $B$ entries of largest $\overline{q}$ \Comment{width; anything dropped stays visible through $u_m$}
  \State move finished entries of $C$ into $\Hyp$; \; $\mathrm{live}\gets C\setminus\Hyp$
  \State $\underline{q}^\ast \gets \max_{h\in\Hyp}\ \min_m p_m(h\mid x)$ \Comment{best lower mass so far}
\EndFor
\Statex
\Statex \textbf{Depth two: Credal Semantic Commitment}
\State merge sequences of $\Hyp$ that are identical after lower-casing and removing punctuation and articles into candidates $a$, with $p_m(a)\gets\sum_{h\in a}p_m(h\mid x)$
\State $u_m \gets 1-\sum_a p_m(a)$ for every $m$ \Comment{unexplored mass, \cref{eq:unexplored}}
\State $e_m(a,b)\gets$ adapter $m$'s probability of ``yes'' when asked whether $a$ and $b$ answer $x$ alike
\State $\Frame(x)\gets$ greedily group the candidates, in decreasing mass, joining $a$ to the meaning whose representative $b$ maximises $\min_m e_m(a,b)$ when that value is $\geq\tfrac12$ \Comment{the adapters judge meaning}
\State $\Plow(c)\gets\min_m \sum_{a\in c}p_m(a)$, \; $\Pup(c)\gets\max_m\bigl[\sum_{a\in c}p_m(a)+u_m\bigr]$ \Comment{\cref{eq:meaningbounds}}
\State $\Dset(\Frame(x))\gets\Call{Dominance}{\Plow,\Pup}$
\If{$|\Dset(\Frame(x))|=1$ \textbf{and} $\Plow(c^\ast)\geq\max_m u_m$} \Comment{second condition: no unseen meaning could dominate}
  \State commit to $c^\ast$ with confidence $\Plow(c^\ast)$
\Else
  \State present $\Dset(\Frame(x))$, with an unseen element when the second condition fails
\EndIf
\end{algorithmic}
\end{algorithm}

\subsection{Worked Example Through Both Depths}
\label{app:walkthrough}

This section follows one prompt through the whole method, drawn step by step below, with the numbers taken from the stored caches of the Llama-3.1-8B run on the CoQA test split ($M{=}5$, $B{=}8$, $L{=}32$), so that every quantity defined in \S\ref{sec:method} can be seen on a real input. The passage describes Diwali and ends with the sentence that families prepare special holiday meals to celebrate the event. The question is \emph{What do they make for the event?} and the gold answer is \emph{special holiday meals}.

\textbf{Step 1: the adapters disagree about the first answer token.} One forward pass of the five adapters gives a distribution over the vocabulary at the first answer position. \Cref{tab:walk_tokens} lists the six most likely tokens with the probability each adapter assigns. Three adapters put most of their mass on the lower-case token \texttt{special}, two prefer the capitalised \texttt{Special}, and the spread on \texttt{special} alone runs from $0.029$ to $0.938$. The bounds of \cref{eq:bounds} are the column-wise minimum and maximum, so $\Plow(\texttt{special})=0.029$ and $\Pup(\texttt{special})=0.938$.

\begin{center}
\begin{tikzpicture}[x=9cm,y=0.55cm,font=\scriptsize]
\draw[black!40] (0,0)--(1,0); \foreach \v in {0,0.5,1}{\draw[black!40](\v,0)--(\v,-0.12) node[below]{$\v$};}
\node[anchor=north] at (0.5,-0.7) {probability of the first answer token under each adapter};
\foreach \tok/\y/\lo/\hi/\vals in {\texttt{special}/2.4/0.029/0.938/{0.684/a,0.029/a,0.245/a,0.925/b,0.938/a},\texttt{Special}/1.1/0.013/0.847/{0.222/a,0.847/a,0.666/a,0.046/b,0.013/a}}{
  \node[anchor=east] at (-0.01,\y) {\tok};
  \draw[blue!25,line width=5pt] (\lo,\y)--(\hi,\y);
  \foreach \v/\pos [count=\m] in \vals {\fill[blue!60!black] (\v,\y) circle (1.6pt);
    \if\pos a\node[above=1pt,font=\tiny] at (\v,\y) {$f_{\m}$};\else\node[below=1pt,font=\tiny] at (\v,\y) {$f_{\m}$};\fi}
}
\end{tikzpicture}
\end{center}

\textbf{Step 2: the rule in the token space.} The largest lower probability over all tokens is $\max_b \Plow(b)=0.029$, attained by \texttt{special}. By \cref{eq:dominance} a token survives when its upper probability reaches that value. Four tokens do: \texttt{special}, \texttt{Special}, \texttt{food} and \texttt{They}, with upper probabilities $0.938$, $0.847$, $0.044$ and $0.037$. The two remaining tokens in \cref{tab:walk_tokens} have upper probabilities $0.018$ and $0.014$ and are ruled out. Since $|\Dset_1|=4$, Credal Token Commitment does not commit and would present the four surviving tokens. This is the failure mode the token space is prone to: the adapters agree about the answer and disagree about how to spell its first token.

\begin{center}
\begin{tikzpicture}[x=2.6cm,y=0.42cm,font=\scriptsize]
\draw[black!40] (0,0)--(2,0);
\foreach \v/\l in {0/0.01,1/0.1,2/1}{\draw[black!40](\v,0)--(\v,-0.2) node[below]{$\l$};}
\node[anchor=north] at (1,-0.9) {upper probability $\Pup$ (log scale)};
\foreach \tok/\y/\u/\k in {\texttt{special}/6/0.938/1,\texttt{Special}/5/0.847/1,\texttt{food}/4/0.044/1,\texttt{They}/3/0.037/1,fifth token/2/0.018/0,sixth token/1/0.014/0}{
  \node[anchor=east] at (-0.03,\y) {\tok};
  \pgfmathsetmacro{\xx}{log10(\u)+2}
  \ifnum\k=1 \fill[blue!60!black] (0,\y-0.3) rectangle (\xx,\y+0.3); \else \fill[black!35] (0,\y-0.3) rectangle (\xx,\y+0.3); \fi
  \node[anchor=west] at (2.08,\y) {$\u$};
}
\pgfmathsetmacro{\th}{log10(0.029)+2}
\draw[red!70!black,dashed,thick] (\th,0.4)--(\th,6.8) node[above,align=center]{$\max_b\Plow(b)=0.029$};
\end{tikzpicture}
\end{center}

\textbf{Step 3: credal decoding.} Starting from the empty partial sequence, the adapters are run one step at a time and every partial sequence that no adapter rules out is expanded. The search takes $34$ decoding steps, of which $17$ expand more than one continuation, and hits the width $B{=}8$ twice, which is recorded as a truncation. It ends with nine completed sequences. After the surface normalisation of \S\ref{sec:method:depths}, which lower-cases and removes punctuation and articles, those nine become five candidates, each carrying the probability every adapter assigns it. The probabilities of the candidates do not sum to one for any adapter, and the shortfall is the unexplored mass $u_m$ of \cref{eq:unexplored}, which ranges from $0.128$ to $0.224$ across the five adapters.

\begin{center}
\begin{tikzpicture}[font=\scriptsize, box/.style={draw=black!50,rounded corners=2pt,align=center,inner sep=3pt,minimum height=0.9cm}, arr/.style={-{Stealth[length=4pt]},black!60,thick}]
\node[box] (a) {empty\\answer};
\node[box,right=0.35cm of a] (b) {expand undominated\\partial answers\\$34$ steps, $17$ branching};
\node[box,right=0.35cm of b] (c) {$9$ completed\\sequences};
\node[box,right=1.3cm of c] (d) {$5$ candidates\\$u_m\in[0.128,\,0.224]$};
\draw[arr] (a)--(b); \draw[arr] (b)--(c); \draw[arr] (c)--node[above]{normalise}(d);
\node[below=0.1cm of b,black!60] {width $B{=}8$ reached twice};
\end{tikzpicture}
\end{center}

\textbf{Step 4: the adapters group the candidates into meanings.} Candidates are judged in decreasing order of mean mass, so \emph{special food} is the representative, and each remaining candidate is judged against it by all five adapters. The five judgements for each of the four pairs are all above one half, the lowest being $0.776$, so every adapter agrees that the five candidates say the same thing. The surely-equivalent relation therefore has a single class and the semantic space is $\Frame(x)=\{c_1\}$ with $c_1$ containing all five candidates.

\begin{center}
\begin{tikzpicture}[font=\scriptsize, cand/.style={draw=orange!70!black,fill=orange!8,rounded corners=2pt,minimum width=3.9cm,inner sep=2pt}, arr/.style={-{Stealth[length=4pt]},black!60}]
\foreach \t [count=\i] in {special food (representative),special holiday meals,they prepare special food,special food and holiday meals,food}{
  \node[cand] (c\i) at (0,-0.5*\i) {\t};}
\node[draw=orange!80!black,fill=orange!20,rounded corners=2pt,inner sep=5pt,align=center] (m) at (5.6,-1.5) {one meaning $c_1$};
\foreach \i in {2,...,5}{\draw[arr] (c\i.east)--(m.west);}
\draw[arr] (c1.east)--(m.west);
\node[align=center,black!70] at (5.6,-2.45) {every adapter judges each pair\\equivalent, lowest $e_m=0.776$};
\end{tikzpicture}
\end{center}

\textbf{Step 5: the rule in the semantic space.} Adapter $m$'s mass on $c_1$ is the sum of its candidate probabilities, which gives $0.855$, $0.785$, $0.872$, $0.776$ and $0.797$ for the five adapters. By \cref{eq:meaningbounds}, $\Plow(c_1)=0.776$ and $\Pup(c_1)=\max_m [P_m(c_1)+u_m]=1.000$. The commitment set is $\Dset(\Frame(x))=\{c_1\}$, and the second condition holds as well, because $\Plow(c_1)=0.776$ is at least the largest unexplored mass $\max_m u_m = 0.224$, so no meaning the search never reached could still dominate. The model is semantically committed to $c_1$ with confidence $0.776$.

\begin{center}
\begin{tikzpicture}[x=9cm,y=0.55cm,font=\scriptsize]
\draw[black!40] (0,0)--(1,0); \foreach \v in {0,0.5,1}{\draw[black!40](\v,0)--(\v,-0.12) node[below]{$\v$};}
\node[anchor=north] at (0.5,-0.7) {probability of the meaning $c_1$};
\node[anchor=east] at (-0.01,1.2) {$c_1$};
\draw[orange!40,line width=5pt] (0.776,1.2)--(1.0,1.2);
\foreach \v [count=\m] in {0.855,0.785,0.872,0.776,0.797} {\fill[orange!80!black] (\v,1.2) circle (1.6pt);}
\node[below=3pt,anchor=north east] at (0.80,1.2) {$\Plow(c_1)=0.776$};
\node[above=3pt,anchor=south east] at (1.0,1.2) {$\Pup(c_1)=1.000$};
\draw[black!60,dashed,thick] (0.224,0.2)--(0.224,2.2) node[above]{$\max_m u_m=0.224$};
\node[green!40!black] at (0.5,-2.1) {singleton set and $0.776\geq0.224$: commit to $c_1$ with confidence $0.776$};
\end{tikzpicture}
\end{center}

\textbf{Step 6: was it right?} The gold answer \emph{special holiday meals} is one of the five candidates and lies in the committed meaning, so the commitment is correct. The example is the common case reported in \S\ref{sec:exp:results}: the token space is defeated by a capitalisation split that has nothing to do with the answer, the semantic space is not, and the confidence attached to the decision is the mass that survives the most pessimistic adapter rather than an average over them.

\begin{table}[h]
\centering
\small
\caption{\textbf{The token space of the worked example.} Probability of each of the six most likely first answer tokens under each of the five adapters, with the bounds of \cref{eq:bounds}. The largest lower probability is $0.029$, so every token whose upper probability reaches $0.029$ survives the rule of \cref{eq:dominance}. Four tokens survive, so the token space does not commit.}
\label{tab:walk_tokens}
\begin{tabular}{lccccc cc c}
\toprule
Token & $f_1$ & $f_2$ & $f_3$ & $f_4$ & $f_5$ & $\Plow$ & $\Pup$ & In $\Dset_1$ \\
\midrule
\texttt{special} & 0.684 & 0.029 & 0.245 & 0.925 & 0.938 & 0.029 & 0.938 & yes \\
\texttt{Special} & 0.222 & 0.847 & 0.666 & 0.046 & 0.013 & 0.013 & 0.847 & yes \\
\texttt{food}    & 0.044 & 0.005 & 0.016 & 0.022 & 0.036 & 0.005 & 0.044 & yes \\
\texttt{They}    & 0.001 & 0.037 & 0.002 & 0.000 & 0.000 & 0.000 & 0.037 & yes \\
fifth token      & 0.018 & 0.003 & 0.002 & 0.004 & 0.006 & 0.002 & 0.018 & no \\
sixth token      & 0.004 & 0.014 & 0.002 & 0.000 & 0.000 & 0.000 & 0.014 & no \\
\bottomrule
\end{tabular}
\end{table}

\section{Implementation Details}
\label{app:implementation}
In this section, we describe the backbones, adapters and inference (\S\ref{app:impl:models}), the meaning-equivalence procedure (\S\ref{app:impl:meaning}), the baseline implementations (\S\ref{app:impl:baselines}), the experimental setup (\S\ref{app:setup}), and the metrics pipeline and compute (\S\ref{app:impl:pipeline}).

\subsection{Backbones, Adapters and Inference}
\label{app:impl:models}
\textbf{Backbones.} Gemma-2-9B-Instruct, Llama-3.1-8B-Instruct, Qwen2.5-7B-Instruct, all frozen.

\textbf{LoRA training.} $M{=}5$ adapters per (backbone, dataset), trained for one epoch on a bootstrap resample of the dataset's training split, batch size $2$ with gradient accumulation $4$, maximum sequence length $1024$, $\alpha{=}16$, dropout $0.1$. The adapters are deliberately heterogeneous, with per-adapter (target modules, rank, learning rate, seed):
$m{=}1$: \{q, v\}, $r{=}4$, $10^{-4}$, seed $42$;
$m{=}2$: \{q, k, v, o\}, $r{=}8$, $3\times10^{-4}$, seed $43$;
$m{=}3$: \{o, gate, up\}, $r{=}16$, $5\times10^{-4}$, seed $44$;
$m{=}4$: \{q, v, down\}, $r{=}4$, $2\times10^{-4}$, seed $45$;
$m{=}5$: \{k, v, gate\}, $r{=}8$, $10^{-4}$, seed $46$.
The seed controls initialisation, the bootstrap draw and data order. This construction treats the ensemble as a set of alternative fine-tuning decisions rather than as replicas of one recipe, which is what makes the family a sensitivity analysis over choices the data does not adjudicate rather than a set of reruns of the same choice. We did not train seed-only or bootstrap-only ensembles at the same budget, so the paper does not separate the contribution of the heterogeneity from that of the resampling.

\textbf{Inference.} Each adapter is run in evaluation mode (dropout off) with the same chat template. Token-space bounds come from one forward pass per adapter on the prompt. Credal decoding runs one priming pass per adapter over the prompt and then one-token steps over the partial sequences with a per-adapter key-value cache; the partial sequences share the prompt and have equal length, so no padding is involved. The first-step distribution is the token-space distribution of the same prompt, so CTC and CSC read the same bounds at depth one. Against a full re-encoding of each partial sequence, cached probabilities agree to within bf16 precision (at most $0.03$ on any token), which can flip a dominance decision only when two bounds are within that tolerance; prompts longer than $2048$ tokens are truncated and counted. Answers end at the backbone's end-of-turn token or at a newline, whichever comes first.

\subsection{Meaning Equivalence}
\label{app:impl:meaning}
\textbf{Meaning equivalence.} Candidates are judged in decreasing order of mean mass. Each new candidate is judged against the representative (highest-mass member) of every existing meaning, by all adapters in one batched pass, with the query ``Question: $x$. Answer 1: $a$. Answer 2: $b$. Do Answer 1 and Answer 2 give the same answer to the question? Reply with yes or no.'' The judgement $e_m(a,b)=p_m(\text{yes})/[p_m(\text{yes})+p_m(\text{no})]$ is the adapter's first-response-token probability of ``yes'' renormalised against ``no'', each summed over its capitalised and space-prefixed single-token spellings. It joins the meaning whose lower equivalence probability is largest if that value is at least $\tfrac12$, and otherwise starts a new meaning; upper equivalence probabilities of at least $\tfrac12$ link meanings in the coarser partition. This assumes equivalence is transitive within a meaning and costs a few dozen short passes per prompt rather than all pairs. The semantic-entropy baselines use their own clustering: microsoft/deberta-large-mnli bidirectional entailment, as in the original papers~\citep{kuhn2023semantic, farquhar2024detecting}, and a cosine-similarity variant with BAAI/bge-base-en-v1.5 embeddings. The quality of both groupings is reported as the rate of prompts on which one meaning contains both a gold-matching and a non-matching candidate.

\subsection{Baseline Implementations}
\label{app:impl:baselines}
\textbf{Baseline implementations.} \emph{Standard LLM}: the base model with the same chat template and prompt, its first-token predictive entropy and max probability, its greedy answer, and $K{=}16$ samples at temperature $0.8$ with maximum length $L$ and their sequence log-probabilities. \emph{Semantic entropy (cosine)}: the $16$ samples clustered greedily by cosine similarity of BAAI/bge-base-en-v1.5 embeddings at threshold $0.8$ (letters on multiple choice), with the length-normalised, renormalised sequence probabilities of \citet{kuhn2023semantic} as cluster masses; \emph{Semantic entropy (NLI)}: the same samples clustered by bidirectional entailment under microsoft/deberta-large-mnli, a pair sharing a meaning only if each entails the other (argmax, no threshold), following \citet{farquhar2024detecting}. \emph{LoRA Ensemble}: the five adapters' first-token distributions averaged; predictive entropy of the mean, mutual information (mean entropy minus expected entropy) and the variance across adapters of the probability of each adapter's most likely token, as in the original code base. \emph{Bayesian-LoRA (KFAC)} and \emph{Laplace-LoRA (diagonal)}: a Laplace posterior over the LoRA weights of the first adapter of each ensemble (the MAP point), with the Fisher estimated on $200$ training examples of the same dataset: for the diagonal method~\citep{daxberger2021laplace} the sum of squared per-example gradients of the answer-token log-likelihood, for KFAC the Kronecker factors of the bayesian-lora package~\citep{yang2023bayesian} at the first answer position with sampled labels and rank-$10$ low-rank factors for wide layers. The prior precision is $\lambda=P/\lVert\theta_{\mathrm{MAP}}\rVert^2$, the maximum-likelihood prior variance of the adapter weights, so that posterior noise in directions without curvature is at the scale of the weights themselves; we record that with the fixed prior precision of $1$ and temperature $0.05$ used in the earlier version of this work the posterior samples destroy the adapter (near-uniform first-token distributions, noise $25$--$45$ times the weights), and that maximising the Laplace marginal likelihood on a $200$-example Fisher is ill-posed because most LoRA directions have no curvature. Each prompt uses $S{=}20$ posterior samples; the scores are the predictive entropy of the posterior predictive (primary) and the mutual information across samples; the answer is the MAP model's greedy answer.

\subsection{Experimental Setup}
\label{app:setup}

\textbf{Backbones and ensemble size.} The three backbones span the dominant open-weight tokeniser and architecture families at the $7$--$9$B scale. $M{=}5$ is a compute-bounded choice; by \cref{prop:monotone} the commitment sets can only grow with $M$, so all comparisons are at fixed $M$ and the dependence on $M$ is reported once in \S\ref{app:ablation:m}.

\textbf{Development splits.} The width $B{=}8$ and the length $L{=}32$ ($L{=}8$ on multiple choice) are fixed on frozen development splits disjoint from every training and test set (the width sweep is in \S\ref{app:ablation:expansion}\textcolor{blue}{; at $L{=}32$, sequences cut at $L$ are $1.9\%$ of the completion set on CoQA and $11.7\%$ on TriviaQA with the passage, for Llama-3.1-8B}), and the cost $\kappa$ of \S\ref{sec:method:depths} is swept rather than fixed (\cref{tab:cost}): for OpenBookQA, $500$ items from the unused remainder of the training split; for ARC-Challenge, the $299$-item validation split; for CoQA and TriviaQA, $500$ items each from the unused remainder of the training split. Test sets are never read during selection.

\textbf{Datasets.} OpenBookQA is four-way multiple choice with a supporting fact; ARC-Challenge is four-way multiple choice requiring multi-step reasoning; CoQA is conversational QA with short free-form answers; TriviaQA is open-domain factual QA with long-tail entity answers. On the two multiple-choice tasks the token space is the whole vocabulary, as everywhere else, and CTC is the decision: every surviving token is an option letter on $97.2$ to $99.0\%$ of questions without the rule being told that they must be ($92.3$ to $99.1\%$ of surviving tokens, the shortfall concentrated on the few questions with large $\Dset_1$). On CoQA and TriviaQA the token space closes on at most $30\%$ of clean prompts (\cref{tab:corrupt_commit}) and CSC is the decision.

\textbf{Corrupted-context protocol.} Where the answer rests on supplied evidence, corruption replaces or perturbs that evidence and leaves the question intact (the passage in CoQA and TriviaQA, the supporting fact in OpenBookQA), so the question stays answerable in principle but the evidence no longer supports the answer. Where there is no evidence (closed-book TriviaQA and ARC-Challenge), the question is replaced and the answer side kept. The four-option test sets have $500$ questions each. Each open-ended setting evaluates $250$ clean and $250$ corrupted prompts, and the paired OpenBookQA split the same $500$ questions with and without corruption; a prompt is labelled corrupted if its evidence or question was replaced.

\textbf{Correctness rules.} Multiple choice: the decoded option letter equals the gold letter. Free-form answers: after SQuAD normalisation (lower case, punctuation and articles removed), the answer contains a gold alias as a token sequence, an alias contains the answer, or token F1 with an alias is at least $0.5$; TriviaQA uses all accepted aliases of the public release, CoQA its single gold string. A presented set is correct if it contains the gold answer.

\subsection{Metrics Pipeline and Compute}
\label{app:impl:pipeline}
\textbf{Metrics pipeline and provenance.} All numbers are produced by one script that reads cached adapter outputs, applies the definitions of \cref{sec:method} once each, and writes a manifest mapping every table cell to its cache file, checksum, code commit and definition. Token-space results on multiple choice use the first-token bounds of the $M{=}5$ ensembles; semantic-space results come from credal decoding, with $M$ stated per row. Unit tests on hand-computed examples cover the bounds, the commitment rule, credal decoding on a mock ensemble (including the width and the cached path), calibration and selective metrics.

\textbf{Compute.} Adapters were trained on single A100 nodes; credal decoding, the equivalence judgements and the baselines ran on a single L4 (24\,GB) or A100 (40\,GB) GPU. Per-stage wall-clock and memory are given in \S\ref{app:cost}.

\section{Additional Experimental Results}
\label{app:additional}
\label{app:results}
\label{app:experiments}
\label{tab:experiment_list}
In this section, we address the following questions based on our experimental findings:
\begin{itemize}[leftmargin=*,topsep=2pt,itemsep=2pt]
\item \textbf{Q1:} How accurate is CLLM at a fixed coverage, and how well calibrated is its confidence, against every baseline family on every backbone and dataset? (\S\ref{app:results:comparison})
\item \textbf{Q2:} Do the credal quantities detect the model's own wrong answers and corrupted context on every backbone, and does the detection depend on which scalar is read off the credal set? (\S\ref{app:results:detection})
\item \textbf{Q3:} Does the rule's own operating point sit where a tuned threshold would, and does the rule withdraw under corrupted context without being tuned? (\S\ref{app:results:selective})
\item \textbf{Q4:} Does the credal set contain the gold answer more often than a top-$k$ set, and what does the rule add to the ranking of the same adapters? (\S\ref{app:results:sets})
\item \textbf{Q5:} How does Credal Token Commitment behave on the four-option setting, and are its intervals calibrated without a calibration step? (\S\ref{app:results:mcq})
\item \textbf{Q6:} What does the semantic space add over the token space, and when does each depth commit? (\S\ref{app:results:depths})
\item \textbf{Q7:} How sensitive are the commitment sets to the choices inside the method, namely the number of adapters, the expansion rule, the width, the unexplored mass and the grouping of candidates into meanings? (\S\ref{app:ablations})
\item \textbf{Q8:} How does commitment behave when a deployment declares the cost of a wrong answer? (\S\ref{app:ablation:cost_sensitive})
\item \textbf{Q9:} What does credal commitment cost at inference? (\S\ref{app:cost})
\end{itemize}

%
\subsection{Selective Accuracy and Calibration per Backbone}
\label{app:results:comparison}
\textbf{Why selective accuracy and calibration come first.} The claim of the paper is about decisions, not about a new score, so the first question is whether the decision the rule takes is at least as good as what a practitioner would get from the baselines they already use. That requires measures every baseline family can be asked for, which is why the comparison reports selective accuracy at a fixed coverage and the calibration of each method's own confidence, on the same prompts for every method; detection and set-valued prediction follow in \S\ref{app:results:detection} and \S\ref{app:results:sets}.

\textbf{Why these baselines.} The Standard LLM is what a practitioner runs by default. The LoRA Ensemble is the same five adapters mean-pooled, which is the closest possible comparison, since any gap is attributable to the reading of the ensemble rather than to the ensemble itself. Bayesian-LoRA and Laplace-LoRA represent posterior approximations over the adapter weights rather than an explicit set of predictors. Semantic entropy with cosine clustering and with entailment clustering represents the sampling-based account of meaning, and both variants are included because the clustering choice, not the entropy, is what drives that family's numbers.

\Cref{tab:comparison_llama} and \cref{tab:comparison_qwen} repeat \cref{tab:comparison} on the other two backbones, reporting accuracy at $80\%$ coverage together with the ECE and NLL of each method's own confidence, and \cref{tab:mcq_selective} reports accuracy at $80\%$ coverage on the four-option setting for the Standard LLM, the LoRA Ensemble and CLLM.

On Llama-3.1-8B (\cref{tab:comparison_llama}), CLLM has the highest accuracy at $80\%$ coverage on CoQA ($0.845$), closed-book TriviaQA ($0.865$), OpenBookQA ($0.980$) and ARC-Challenge ($0.890$), and the second highest on TriviaQA with the passage ($0.920$ against $0.940$ for LoRA Ensemble (predictive entropy)). On CoQA the single-model and posterior baselines reach $0.745$--$0.775$. On Qwen2.5-7B (\cref{tab:comparison_qwen}), CLLM has the highest accuracy at $80\%$ coverage on CoQA ($0.825$) and OpenBookQA ($0.980$), is $0.002$ behind the best on ARC-Challenge, and ties Semantic entropy (NLI) for second on closed-book TriviaQA ($0.620$ against $0.660$), where it improves over the Standard LLM's $0.545$. On Qwen its confidence is the best calibrated of all methods on both four-option datasets, with ECE $0.039$ and $0.066$ and NLL $0.209$ and $0.327$ on OpenBookQA and ARC-Challenge, and it has the lowest ECE on CoQA ($0.114$) and the lowest NLL on closed-book TriviaQA ($0.535$). Averaged over CoQA and TriviaQA with the passage, CLLM is above every single-model baseline in accuracy at $80\%$ coverage on all three backbones, its sets contain the gold meaning on $89$, $89$ and $88\%$ of questions (\cref{tab:csc}), and it is the best corrupted-context detector on Llama and Gemma ($0.855$ and $0.878$). On the four-option setting (\cref{tab:mcq_selective}), the token-space lower bound is the best ranker on both datasets for Llama and Gemma and on OpenBookQA for Qwen; on ARC-Challenge with Qwen it ties the Standard LLM at $0.938$, $0.002$ behind the LoRA Ensemble, and it is at least as accurate as the Standard LLM in every cell.

\begin{table*}[h]
\centering
\setlength{\tabcolsep}{3pt}
\caption{\textbf{Comparison with baselines on every dataset, Llama-3.1-8B} (CoQA, TriviaQA with and without the passage, OpenBookQA and ARC-Challenge). Each dataset carries three measures: \emph{Acc.@80\%}, the accuracy of the selected answer when the questions are ranked by the method's score and the easiest $80\%$ are answered, higher is better; \emph{ECE} and \emph{NLL}, the expected calibration error and negative log likelihood of the method's own per-answer confidence, lower is better in both. For the credal row the confidence is the midpoint of the interval. Rows that select the same answer share that answer's confidence, so the two Standard LLM rows and the two LoRA Ensemble rows carry the same calibration. CLLM is one row, at the depth the answer type fixes. \textbf{Bold} = best per column, \underline{underlined} = second-best; the grey row is ours.}
\label{tab:comparison_llama}
\resizebox{\textwidth}{!}{\begin{tabular}{lccccccccccccccc}
\toprule
& \multicolumn{3}{c}{CoQA} & \multicolumn{3}{c}{TriviaQA (passage)} & \multicolumn{3}{c}{TriviaQA (closed book)} & \multicolumn{3}{c}{OpenBookQA} & \multicolumn{3}{c}{ARC-Challenge} \\
\cmidrule(lr){2-4} \cmidrule(lr){5-7} \cmidrule(lr){8-10} \cmidrule(lr){11-13} \cmidrule(lr){14-16}
Score & Acc.@80\%~$\uparrow$ & ECE~$\downarrow$ & NLL~$\downarrow$ & Acc.@80\%~$\uparrow$ & ECE~$\downarrow$ & NLL~$\downarrow$ & Acc.@80\%~$\uparrow$ & ECE~$\downarrow$ & NLL~$\downarrow$ & Acc.@80\%~$\uparrow$ & ECE~$\downarrow$ & NLL~$\downarrow$ & Acc.@80\%~$\uparrow$ & ECE~$\downarrow$ & NLL~$\downarrow$ \\
\midrule
Standard LLM (predictive entropy) & 0.775 & 0.083 & 0.511 & 0.895 & 0.075 & \textbf{0.342} & 0.785 & 0.117 & 0.563 & 0.965 & 0.040 & 0.240 & 0.858 & 0.104 & 0.504 \\
Standard LLM (max prob) & 0.770 & 0.083 & 0.511 & 0.890 & 0.075 & \textbf{0.342} & 0.785 & 0.117 & 0.563 & \underline{0.968} & 0.040 & 0.240 & \underline{0.860} & 0.104 & 0.504 \\
\midrule
Semantic entropy (cosine) & 0.755 & 0.154 & 1.825 & 0.885 & \underline{0.070} & 0.765 & 0.795 & 0.136 & 1.127 & \underline{0.968} & 0.047 & 0.622 & \underline{0.860} & 0.131 & 1.458 \\
Semantic entropy (NLI) & 0.745 & 0.177 & 0.871 & 0.905 & {0.291} & 0.565 & 0.810 & 0.229 & 0.829 & 0.945 & 0.175 & 0.447 & \underline{0.860} & \textbf{0.041} & 0.821 \\
\midrule
Bayesian-LoRA (KFAC) & 0.745 & \underline{0.062} & \underline{0.510} & 0.890 & 0.438 & 0.855 & 0.785 & 0.103 & 0.543 & 0.963 & {0.025} & 0.224 & 0.855 & 0.073 & \underline{0.415} \\
Laplace-LoRA (diagonal) & 0.750 & {0.089} & {0.816} & 0.890 & 0.442 & 0.860 & 0.780 & 0.107 & 0.537 & 0.963 & 0.026 & 0.223 & \underline{0.860} & 0.071 & \textbf{0.414} \\
\midrule
LoRA Ensemble (predictive entropy) & \underline{0.840} & 0.073 & 0.845 & \textbf{0.940} & 0.148 & 0.432 & \underline{0.850} & \underline{0.100} & \textbf{0.429} & 0.965 & \underline{0.024} & \textbf{0.182} & {0.810} & \underline{0.053} & 0.426 \\
LoRA Ensemble (mutual information) & 0.815 & 0.073 & 0.845 & 0.875 & 0.148 & 0.432 & 0.795 & {0.100} & \textbf{0.429} & \underline{0.968} & \underline{0.024} & \textbf{0.182} & 0.853 & \underline{0.053} & 0.426 \\
\midrule
\rowcolor{gray!15} \textbf{CLLM} & \textbf{0.845} & \textbf{0.053} & \textbf{0.493}
 & \underline{0.920} & \textbf{0.068} & \underline{0.420} & \textbf{0.865} & \textbf{0.050} & 0.461 & \textbf{0.980} & \textbf{0.018} & \underline{0.185} & \textbf{0.890} & 0.069 & 0.419 \\
\bottomrule
\end{tabular}
}
\end{table*}

\begin{table*}[h]
\centering
\setlength{\tabcolsep}{3pt}
\caption{\textbf{Comparison with baselines on every dataset, Qwen2.5-7B} (CoQA, TriviaQA with and without the passage, OpenBookQA and ARC-Challenge). Each dataset carries three measures: \emph{Acc.@80\%}, the accuracy of the selected answer when the questions are ranked by the method's score and the easiest $80\%$ are answered, higher is better; \emph{ECE} and \emph{NLL}, the expected calibration error and negative log likelihood of the method's own per-answer confidence, lower is better in both. For the credal row the confidence is the midpoint of the interval. Rows that select the same answer share that answer's confidence, so the two Standard LLM rows and the two LoRA Ensemble rows carry the same calibration. CLLM is one row, at the depth the answer type fixes. \textbf{Bold} = best per column, \underline{underlined} = second-best; the grey row is ours.}
\label{tab:comparison_qwen}
\resizebox{\textwidth}{!}{\begin{tabular}{lccccccccccccccc}
\toprule
& \multicolumn{3}{c}{CoQA} & \multicolumn{3}{c}{TriviaQA (passage)} & \multicolumn{3}{c}{TriviaQA (closed book)} & \multicolumn{3}{c}{OpenBookQA} & \multicolumn{3}{c}{ARC-Challenge} \\
\cmidrule(lr){2-4} \cmidrule(lr){5-7} \cmidrule(lr){8-10} \cmidrule(lr){11-13} \cmidrule(lr){14-16}
Score & Acc.@80\%~$\uparrow$ & ECE~$\downarrow$ & NLL~$\downarrow$ & Acc.@80\%~$\uparrow$ & ECE~$\downarrow$ & NLL~$\downarrow$ & Acc.@80\%~$\uparrow$ & ECE~$\downarrow$ & NLL~$\downarrow$ & Acc.@80\%~$\uparrow$ & ECE~$\downarrow$ & NLL~$\downarrow$ & Acc.@80\%~$\uparrow$ & ECE~$\downarrow$ & NLL~$\downarrow$ \\
\midrule
Standard LLM (predictive entropy) & 0.740 & 0.212 & 1.173 & 0.735 & 0.225 & 1.021 & 0.545 & 0.370 & 2.418 & 0.975 & 0.090 & 0.940 & \underline{0.938} & 0.109 & 1.459 \\
Standard LLM (max prob) & 0.740 & 0.212 & 1.173 & 0.715 & 0.225 & 1.021 & 0.530 & 0.370 & 2.418 & 0.975 & 0.090 & 0.940 & \underline{0.938} & 0.109 & 1.459 \\
\midrule
Semantic entropy (cosine) & 0.745 & 0.207 & 4.255 & 0.750 & 0.208 & 3.084 & 0.585 & 0.371 & 6.550 & 0.920 & 0.091 & 1.752 & 0.897 & 0.113 & 2.341 \\
Semantic entropy (NLI) & 0.770 & 0.145 & 2.213 & 0.790 & 0.100 & 1.252 & \underline{0.620} & 0.130 & 1.003 & 0.900 & 0.138 & 1.645 & 0.885 & 0.131 & 1.331 \\
\midrule
Bayesian-LoRA (KFAC) & 0.770 & 0.128 & \underline{0.665} & 0.845 & 0.067 & 0.485 & 0.555 & 0.225 & 1.119 & 0.975 & 0.085 & 0.831 & 0.935 & 0.106 & 1.392 \\
Laplace-LoRA (diagonal) & 0.770 & 0.118 & \textbf{0.641} & 0.845 & 0.062 & 0.473 & 0.555 & 0.212 & 1.067 & 0.975 & 0.080 & 0.767 & 0.935 & 0.104 & 1.284 \\
\midrule
LoRA Ensemble (predictive entropy) & {0.743} & {0.126} & 0.894 & \textbf{0.905} & \textbf{0.047} & \textbf{0.376} & \textbf{0.660} & \textbf{0.089} & \underline{0.566} & \underline{0.978} & \underline{0.058} & \underline{0.235} & \textbf{0.940} & \underline{0.080} & \underline{0.362} \\
LoRA Ensemble (mutual information) & \underline{0.790} & {0.126} & 0.894 & 0.840 & \textbf{0.047} & \textbf{0.376} & 0.595 & \textbf{0.089} & \underline{0.566} & \underline{0.978} & \underline{0.058} & \underline{0.235} & 0.930 & \underline{0.080} & \underline{0.362} \\
\midrule
\rowcolor{gray!15} \textbf{CLLM} & \textbf{{0.825}} & \textbf{0.114} & 0.906 & \underline{0.860} & 0.092 & 0.443 & \underline{0.620} & 0.130 & \textbf{0.535} & \textbf{0.980} & \textbf{0.039} & \textbf{0.209} & \underline{0.938} & \textbf{0.066} & \textbf{0.327} \\
\bottomrule
\end{tabular}
}
\end{table*}

\begin{table}[h]\centering\small
\caption{Selective accuracy at $80\%$ coverage on the four-option setting, OpenBookQA and ARC-Challenge ($N{=}500$; ensemble scores rank the representative answer, the Standard LLM ranks its greedy answer).}
\label{tab:mcq_selective}
\resizebox{\columnwidth}{!}{\begin{tabular}{llccc}
\toprule
\textbf{Dataset} & \textbf{Backbone} & Standard LLM (predictive entropy)~$\uparrow$ & LoRA Ensemble (predictive entropy)~$\uparrow$ & \textbf{CLLM}~$\uparrow$ \\
\midrule
OpenBookQA & Llama-3.1-8B & \underline{0.965} & \underline{0.965} & \cellcolor{gray!15}\textbf{0.980} \\
OpenBookQA & Gemma-2-9B & 0.973 & \underline{0.995} & \cellcolor{gray!15}\textbf{0.998} \\
OpenBookQA & Qwen2.5-7B & 0.975 & \underline{0.978} & \cellcolor{gray!15}\textbf{0.980} \\
ARC-Challenge & Llama-3.1-8B & \underline{0.858} & 0.810 & \cellcolor{gray!15}\textbf{0.890} \\
ARC-Challenge & Gemma-2-9B & 0.943 & \underline{0.950} & \cellcolor{gray!15}\textbf{0.953} \\
ARC-Challenge & Qwen2.5-7B & \underline{0.938} & \textbf{0.940} & \cellcolor{gray!15}\underline{0.938} \\
\bottomrule
\end{tabular}
}
\end{table}

\subsection{Hallucination and Corrupted-Context Detection}
\label{app:results:detection}
\begin{figure*}[h]
\centering
\includegraphics[width=\textwidth]{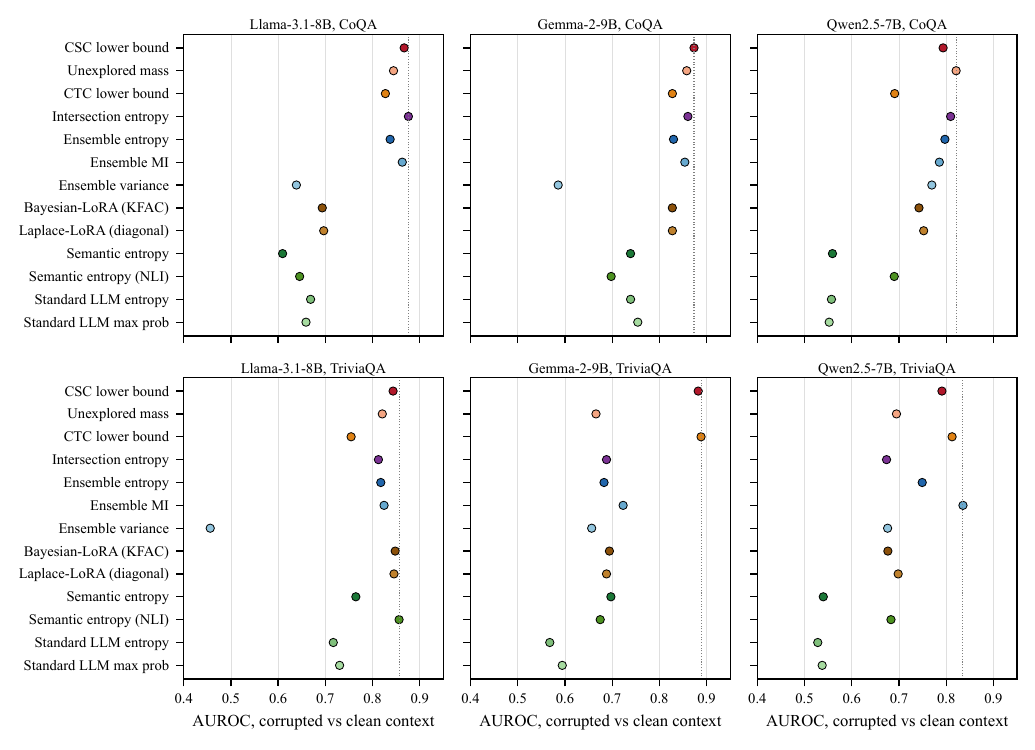}
\caption{\textbf{Corrupted-context detection across all baseline families}, AUROC on $250$ clean against $250$ corrupted prompts per cell. Credal quantities are shown against the mean-pooled ensemble scores, the two posterior approximations and the single-model scores, for both open-ended datasets and all three backbones.}
\label{fig:detection_forest}
\end{figure*}

\Cref{fig:detection_forest} shows the corrupted-context AUROC of every method on CoQA and TriviaQA with the passage for all three backbones. On Llama the CSC lower bound is ahead of every baseline on CoQA ($0.867$) and fourth on TriviaQA ($0.844$), behind semantic entropy (NLI) and the two posteriors. On Qwen it is second on both datasets ($0.793$ and $0.791$), behind the LoRA Ensemble's predictive entropy on CoQA and its mutual information on TriviaQA. On Gemma TriviaQA the token-space lower bound ($0.888$) is ahead of every baseline.

\cref{tab:detection_controls} reports corrupted-versus-clean AUROC for scalar statistics read from the token-space credal set and for the same adapters mean-pooled. The credal statistics are: the entropy of the distribution that interpolates every interval by the same fraction (the intersection-probability transform of \citet{cuzzolin09-intersection}), the upper entropy over the credal set~\citep{wang2024credalwrapper}, the mean interval width, the token-space lower bound $1-\Plow(a^*)$ and the unexplored mass $\max_m u_m(x)$ of \cref{eq:unexplored}. The two set entropies and the width stay within $0.08$ of LoRA Ensemble (predictive entropy) in every cell, and Standard LLM (predictive entropy) is far below both on CoQA and TriviaQA. The ensemble, not the readout, is what separates corrupted from clean context by ranking; the contribution of the credal set is the decision of \cref{sec:method:rule}, not a ranker. The token-space lower bound departs furthest from the mean-pooled entropy, in opposite directions: it falls to $0.691$ on Qwen CoQA against $0.797$, and reaches $0.888$ on Gemma TriviaQA against $0.683$. The unexplored mass is the best statistic in the table on Qwen CoQA at $0.821$, above every method in \cref{fig:detection_forest} for that cell, but falls to $0.666$ on Gemma TriviaQA, so it stays an ablation rather than joining \cref{tab:detection}.

\begin{table*}[!h]
\centering
\setlength{\tabcolsep}{3pt}
\caption{\textbf{Scalar statistics read from the token-space credal set, as corrupted-versus-clean detectors} (AUROC; $250$ clean and $250$ corrupted prompts per cell, $M{=}5$). Every ensemble and credal row is read from the same five adapters. The rows that also appear in \cref{tab:detection} carry the same values there. \textbf{Bold} = best per column, \underline{underlined} = second best; grey rows are ours.}
\label{tab:detection_controls}
\resizebox{\textwidth}{!}{\begin{tabular}{l cccccc}
\toprule
& \multicolumn{3}{c}{\textbf{CoQA}} & \multicolumn{3}{c}{\textbf{TriviaQA (passage)}} \\
\cmidrule(lr){2-4} \cmidrule(lr){5-7}
\textbf{Statistic} (AUROC~$\uparrow$) & Llama & Gemma & Qwen & Llama & Gemma & Qwen \\
\midrule
Single model, predictive entropy & 0.669 & 0.739 & 0.557 & 0.717 & 0.568 & 0.528 \\
Mean-pooled ensemble, predictive entropy & 0.837 & {0.830} & 0.797 & 0.817 & 0.683 & 0.749 \\
Mean-pooled ensemble, mutual information & \underline{0.863} & 0.854 & 0.786 & \textbf{0.824} & \underline{0.723} & \textbf{0.836} \\
\midrule
\rowcolor{gray!15} Credal set, entropy of the interpolating distribution & \textbf{0.876} & \textbf{0.860} & \underline{0.809} & 0.813 & 0.688 & 0.674 \\
\rowcolor{gray!15} Credal set, upper entropy & 0.855 & 0.854 & 0.805 & 0.779 & 0.690 & 0.736 \\
\rowcolor{gray!15} Credal set, mean interval width & \textbf{0.876} & 0.824 & 0.786 & 0.799 & 0.667 & 0.807 \\
\rowcolor{gray!15} Credal set, $1-\Plow(a^*)$ (CTC) & 0.827 & 0.828 & 0.691 & 0.755 & \textbf{0.888} & \underline{0.813} \\
\rowcolor{gray!15} Credal set, unexplored mass $\max_m u_m(x)$ & 0.844 & \underline{0.858} & \textbf{0.821} & \underline{0.821} & 0.666 & 0.695 \\
\bottomrule
\end{tabular}
}
\end{table*}

\subsection{Selective Prediction and Abstention}
\label{app:results:selective}
\textbf{Why the corrupted-context protocol exists.} Abstention is only meaningful if there are inputs on which the model should abstain. Replacing the evidential passage while leaving the question intact creates exactly that: the question remains answerable in principle but the evidence no longer supports any answer. It gives a clean contrast for detection, and it lets abstention be measured at a coverage matched to what the rule chose on clean prompts, which is the only fair way to compare a rule that sets its own coverage with baselines that need a threshold.

\begin{figure*}[h]\centering\includegraphics[width=\textwidth]{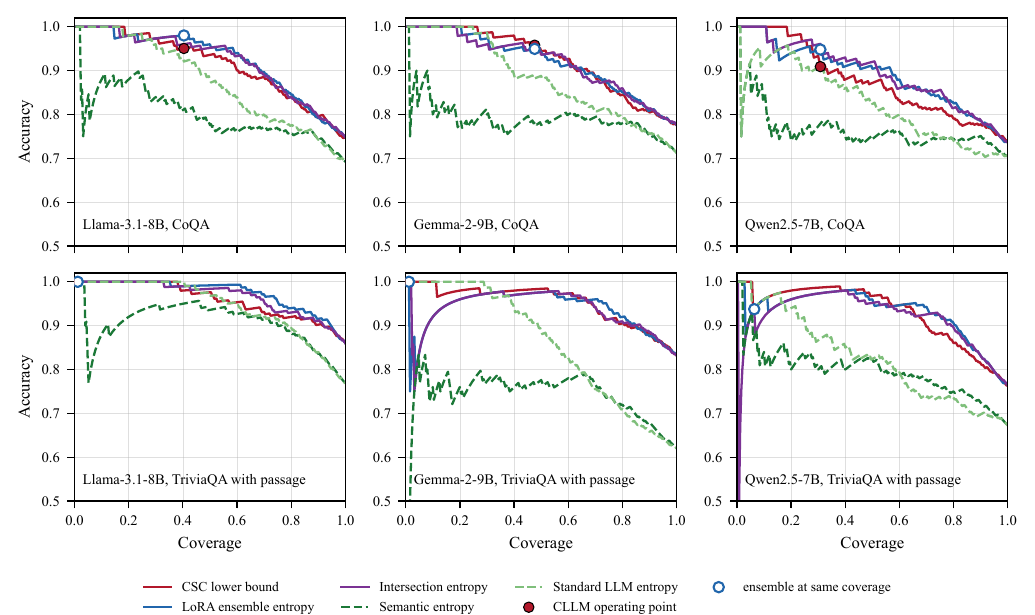}
\caption{\textbf{Risk-coverage curves on clean prompts}, CoQA and TriviaQA with the passage, all three backbones. The horizontal axis is the fraction of questions answered when they are ranked by each method's score, the vertical axis the accuracy of the answers given; a curve that stays high as it moves right ranks its own errors well. The filled marker is the operating point our rule chose for itself, with no threshold; the hollow marker is the LoRA Ensemble at the same coverage.}
\label{fig:risk_coverage_full}
\end{figure*}

\begin{figure*}[h]\centering\includegraphics[width=\textwidth]{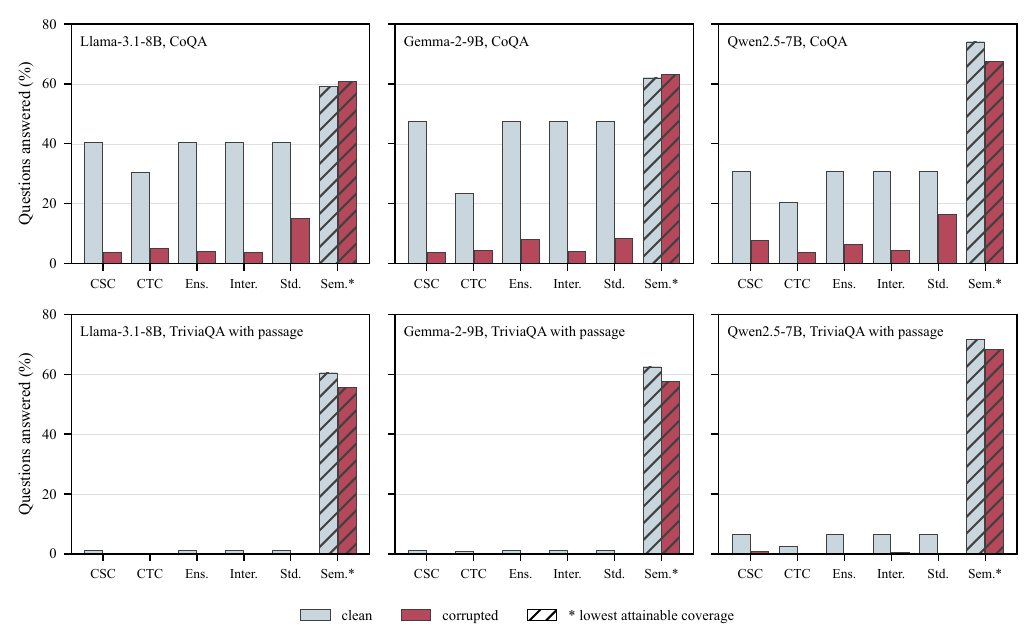}
\caption{\textbf{Abstention at matched clean coverage}, both open-ended datasets, all three backbones. Each method is thresholded on clean prompts to the coverage our rule chose for itself, then shown the corrupted prompts; the bars are the fraction of questions it still answers. A method that answers as much of the corrupted split as of the clean one has not noticed the intervention.}
\label{fig:abstention_full}
\end{figure*}

\Cref{fig:risk_coverage_full} extends the risk-coverage curves of \cref{fig:risk_coverage} to both open-ended datasets and all three backbones. On CoQA, the operating point the rule chose for itself (filled marker) lies on the curves of the same adapters, and the single-model curves, the Standard LLM's entropy and semantic entropy, lie below the ensemble curves over most of the coverage range. On TriviaQA with the passage the rule commits on only $1$--$6\%$ of questions (\cref{tab:csc}), so its operating point sits at the left edge, where it answers with $94$--$100\%$ accuracy. \Cref{fig:abstention_full} thresholds every method on clean prompts to the coverage the rule chose and reports how many prompts each still answers, clean and corrupted. On CoQA the credal rule answers $3.6$--$7.6\%$ of corrupted prompts, whereas semantic entropy, which cannot be thresholded down to the target coverage (hatched), keeps answering most of them; on TriviaQA with the passage every method except semantic entropy answers almost nothing, because the coverage the rule chose is already close to zero.

\subsection{Set-Valued Prediction}
\label{app:results:sets}
The presented meaning sets are compared below with the LoRA Ensemble top-$k$ meanings and with the Standard LLM top-$k$ sampled answers of the same size.

The credal set contains the gold meaning on $86$--$91\%$ of clean questions with the passage and on $76$--$88\%$ of closed-book questions, against $66$--$86\%$ for the Standard LLM's top-$k$ of the same size and $62$--$77\%$ for its greedy answer. Under corrupted context the gap widens: on TriviaQA the credal set still contains the gold answer on $73$--$76\%$ of questions, against $5$--$35\%$ for the Standard LLM's top-$k$. \Cref{tab:setsize} separates the size of the set from its content. Against a $k$ fixed in advance for the whole dataset, the credal set contains the gold meaning more often than the top-$k$ of the same adapters for $k{=}1$ and $k{=}2$ in every cell, and for $k{=}3$ in every cell except closed-book TriviaQA on Llama and Gemma, where the two tie; on CoQA with Llama and Gemma it does so while showing fewer than three meanings on average ($2.52$ and $2.65$). The size the rule chooses per question is therefore what a fixed $k$ cannot provide.

\begin{table}[h]
\centering\footnotesize
\setlength{\tabcolsep}{5pt}
\caption{\textbf{The size, not the ranking} (clean prompts, gold-in-set, per cent). Given our per-question size the ensemble's set is ours on almost every prompt, so those columns agree; against any $k$ fixed in advance the credal set is ahead or level, on CoQA while showing fewer meanings.}
\label{tab:setsize}
\resizebox{\textwidth}{!}{\begin{tabular}{llcccccc}
\toprule
& & & \multicolumn{2}{c}{Matched size} & \multicolumn{3}{c}{Fixed size} \\
\cmidrule(lr){4-5} \cmidrule(lr){6-8}
Dataset & Backbone & Our mean size~$\downarrow$ & Ours~$\uparrow$ & Top-$k$ at our size~$\uparrow$ & $k{=}1$ & $k{=}2$ & $k{=}3$ \\
\midrule
CoQA & Llama-3.1-8B & 2.52 & \textbf{87} & 87 & 75 & 83 & 84 \\
CoQA & Gemma-2-9B & 2.65 & \textbf{90} & 90 & 79 & 87 & 89 \\
CoQA & Qwen2.5-7B & 5.94 & \textbf{89} & 89 & 76 & 82 & 85 \\
TriviaQA (passage) & Llama-3.1-8B & 3.78 & \textbf{91} & 91 & 85 & 88 & 90 \\
TriviaQA (passage) & Gemma-2-9B & 2.51 & \textbf{88} & 88 & 80 & 85 & 87 \\
TriviaQA (passage) & Qwen2.5-7B & 6.44 & \textbf{86} & 86 & 75 & 80 & 83 \\
TriviaQA (closed book) & Llama-3.1-8B & 2.80 & \textbf{88} & 88 & 78 & 83 & 88 \\
TriviaQA (closed book) & Gemma-2-9B & 2.92 & \textbf{86} & 86 & 75 & 82 & 86 \\
TriviaQA (closed book) & Qwen2.5-7B & 7.16 & \textbf{76} & 76 & 56 & 64 & 70 \\
\bottomrule
\end{tabular}
}
\end{table}

\subsection{Four-Option Setting and Interval Calibration}
\label{app:results:mcq}
\textbf{Why the four-option setting is separate.} On multiple choice the answer is one of four letters, so no generation is involved and the token space alone decides. This isolates the commitment rule from credal decoding: any difference on OpenBookQA and ARC-Challenge is attributable to the bounds and the rule, not to the search. ARC-Challenge is answered with the adapters trained on OpenBookQA, which also tests whether the rule survives a shift in the question distribution it was never fine-tuned on.

\textbf{Why interval calibration is measured in that setting.} A point confidence can be checked against accuracy with a reliability diagram, but an interval makes a different promise, namely that it contains the accuracy. The four-option setting is where that promise can be checked cleanly, because correctness is unambiguous and every question has the same four options, so questions can be binned by the midpoint of the selected answer's interval and the bin's accuracy compared with the bin's mean bounds.

\begin{figure*}[h]
\centering
\includegraphics[width=\textwidth]{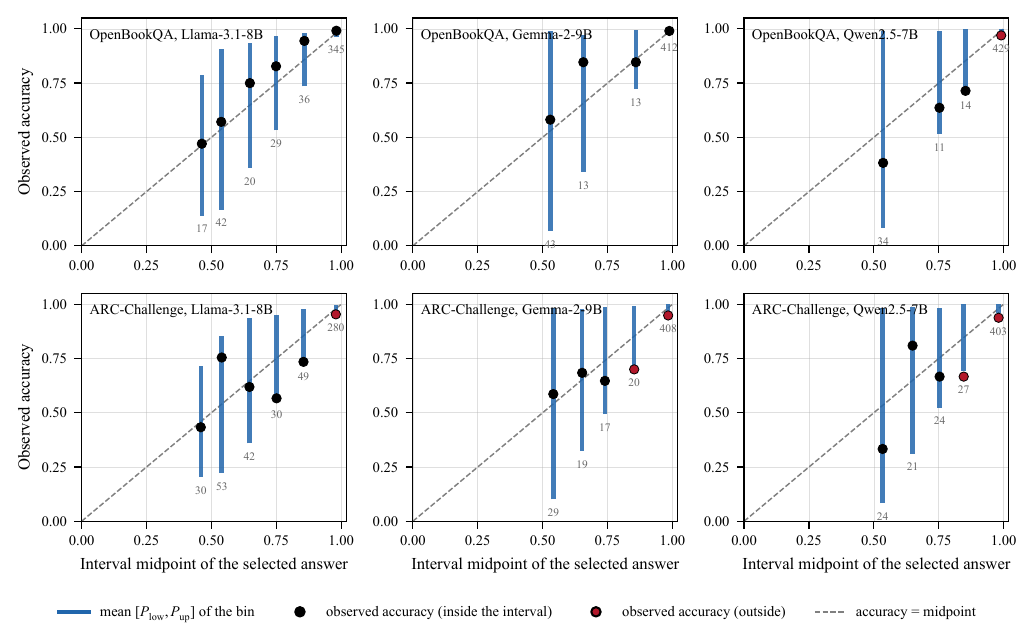}
\caption{\textbf{Interval calibration on the four-option setting.} Questions are binned by the midpoint of the selected answer's interval; the bar is the bin's mean $[\Plow,\Pup]$, the dot the observed accuracy (red when outside the interval), the number the bin count; bins with fewer than $10$ questions are dropped. The table form is \cref{tab:interval_cal}.}
\label{fig:interval_calibration}
\end{figure*}

\Cref{tab:ctc_mcq_full} gives the commit rates behind \cref{fig:commit_space}. On the standard test splits, CTC commits on $73$--$91\%$ of questions and is correct on $89$--$98\%$ of those (\cref{fig:commit_space}). The presented sets contain the gold letter on $89$--$98\%$ of questions at a mean size of $1.14$--$1.50$ letters. On the paired OpenBookQA split, corrupting the passage lowers commitment from $86.8$--$91.6\%$ to $65.2$--$70.6\%$ with no parameter tuned, while the committed answers stay $89$--$95\%$ correct and the sets grow from $1.12$--$1.24$ to $1.77$--$2.21$ letters. \Cref{fig:interval_calibration} and \cref{tab:interval_cal} test the intervals themselves. Across the six cells, the accuracy of a bin lies inside the bin's mean interval in $24$ of the $30$ populated bins ($13$ of $14$ on OpenBookQA and $11$ of $16$ on ARC-Challenge), with no calibration step. The top bin holds $58$--$88\%$ of questions at an accuracy of $0.94$--$0.99$ with intervals $0.02$--$0.04$ wide, and the six bins outside their interval all lie below the lower bound, most of them on ARC-Challenge, which the adapters were not trained on.

\begin{table*}[h]
\centering
\setlength{\tabcolsep}{3pt}
\caption{\textbf{Credal Token Commitment on the four-option setting}, the commit rates behind \cref{fig:commit_space} (OpenBookQA and ARC-Challenge, $N{=}500$, $M{=}5$; ARC-Challenge uses the OpenBookQA-trained adapters). Commit $=|\Dset_1|{=}1$. The final block is the paired OpenBookQA split, the same $500$ questions with and without a corrupted passage, so its clean rows differ slightly from the first block, which uses the standard test split.}
\label{tab:ctc_mcq_full}
\resizebox{\textwidth}{!}{\begin{tabular}{llccc cc}
\toprule
\textbf{Dataset} & \textbf{Backbone} & Commit rate & Mean $|\Dset_1|$~$\downarrow$ & Gold in set~$\uparrow$ & \multicolumn{2}{c}{Acc.\ at the rule's coverage} \\
\cmidrule(lr){6-7}
 & & & & & CTC (ours) & LoRA Ensemble (pred.\ entropy) \\
\midrule
\rowcolor{gray!15} OpenBookQA & Llama-3.1-8B & 0.830 & 1.29 & 0.966 & \textbf{0.971} & \underline{0.959} \\
\rowcolor{gray!15} OpenBookQA & Gemma-2-9B & 0.864 & 1.30 & 0.984 & \underline{0.984} & \textbf{0.986} \\
\rowcolor{gray!15} OpenBookQA & Qwen2.5-7B & 0.906 & 1.17 & 0.956 & \textbf{0.954} & \textbf{0.954} \\
\rowcolor{gray!15} ARC-Challenge & Llama-3.1-8B & 0.734 & 1.50 & 0.888 & \textbf{0.905} & \underline{0.888} \\
\rowcolor{gray!15} ARC-Challenge & Gemma-2-9B & 0.882 & 1.20 & 0.930 & \textbf{0.934} & \underline{0.927} \\
\rowcolor{gray!15} ARC-Challenge & Qwen2.5-7B & 0.892 & 1.14 & 0.912 & \textbf{0.917} & \underline{0.908} \\
\midrule
\multicolumn{7}{l}{\emph{OpenBookQA, the same $500$ questions with and without a corrupted passage}} \\
\rowcolor{gray!15} OpenBookQA, clean & Llama-3.1-8B & 0.868 & 1.24 & 0.952 & \textbf{0.963} & \underline{0.956} \\
\rowcolor{gray!15} OpenBookQA, clean & Gemma-2-9B & 0.898 & 1.22 & 0.986 & \textbf{0.986} & \underline{0.984} \\
\rowcolor{gray!15} OpenBookQA, clean & Qwen2.5-7B & 0.916 & 1.12 & 0.962 & \textbf{0.963} & \underline{0.959} \\
\rowcolor{gray!15} OpenBookQA, corrupted context & Llama-3.1-8B & 0.652 & 1.77 & 0.908 & \textbf{0.890} & \underline{0.880} \\
\rowcolor{gray!15} OpenBookQA, corrupted context & Gemma-2-9B & 0.706 & 2.21 & 0.954 & \textbf{0.963} & \underline{0.952} \\
\rowcolor{gray!15} OpenBookQA, corrupted context & Qwen2.5-7B & 0.670 & 1.92 & 0.928 & \textbf{0.922} & \underline{0.904} \\
\bottomrule
\end{tabular}
}
\end{table*}

\begin{table*}[h]\centering\small
\caption{\textbf{Interval calibration on the four-option setting}, OpenBookQA and ARC-Challenge. All questions binned by the midpoint of the representative answer's interval ($10$ bins; bins with fewer than $10$ questions dropped). A bin is ``inside'' when its accuracy lies within the bin's mean $[\Plow(a^*),\Pup(a^*)]$. The last three columns describe the top bin.}
\label{tab:interval_cal}
\resizebox{\textwidth}{!}{\begin{tabular}{llcccc ccc}
\toprule
& & \multicolumn{4}{c}{\textbf{All populated bins}} & \multicolumn{3}{c}{\textbf{Top bin}} \\
\cmidrule(lr){3-6} \cmidrule(lr){7-9}
\textbf{Dataset} & \textbf{Backbone} & Bins & Inside~$\uparrow$ & Below $\Plow$~$\downarrow$ & Mean width~$\downarrow$ & Share of questions & Accuracy & Mean $[\Plow,\Pup]$ \\
\midrule
\rowcolor{gray!15} OpenBookQA & Llama-3.1-8B & 6 & 6 & 0 & 0.447 & 0.71 & 0.99 & $[0.96,1.00]$ \\
\rowcolor{gray!15} OpenBookQA & Gemma-2-9B & 4 & 4 & 0 & 0.462 & 0.86 & 0.99 & $[0.98,1.00]$ \\
\rowcolor{gray!15} OpenBookQA & Qwen2.5-7B & 4 & 3 & 1 & 0.423 & 0.88 & 0.97 & $[0.98,1.00]$ \\
\rowcolor{gray!15} ARC-Challenge & Llama-3.1-8B & 6 & 5 & 1 & 0.399 & 0.58 & 0.95 & $[0.96,1.00]$ \\
\rowcolor{gray!15} ARC-Challenge & Gemma-2-9B & 5 & 3 & 2 & 0.468 & 0.83 & 0.95 & $[0.97,1.00]$ \\
\rowcolor{gray!15} ARC-Challenge & Qwen2.5-7B & 5 & 3 & 2 & 0.475 & 0.81 & 0.94 & $[0.96,1.00]$ \\
\bottomrule
\end{tabular}
}
\end{table*}

\subsection{Semantic Commitment and Agreement between Depths}
\label{app:results:depths}
\textbf{Why a closed-book setting is included.} With a passage in the prompt, much of the ensemble's agreement comes from the passage rather than from the model. Closed-book TriviaQA removes the passage, so the intervals reflect what the adapters know rather than what they were given, and the semantic space can be compared in both regimes.

\Cref{tab:csc} reports the semantic space. CSC commits on $30.8$--$47.6\%$ of clean CoQA questions and on $24.4$--$45.6\%$ of closed-book TriviaQA questions, with an accuracy when committed of $91$--$96\%$, against $56$--$78\%$ for the representative answer on all questions. Corrupting the passage lowers commitment to $3.6$--$7.6\%$ on CoQA and to at most $0.8\%$ on TriviaQA, and the unexplored mass rises from $0.36$--$0.51$ to $0.73$--$0.88$ on CoQA and from $0.78$--$0.80$ to $0.86$--$0.96$ on TriviaQA: with the passage replaced, the adapters spread their mass over completions the search cannot reach, and the widened upper bounds block commitment.

\begin{table}[h]\centering\footnotesize\setlength{\tabcolsep}{4pt}
\caption{\textbf{Credal Semantic Commitment} on CoQA and TriviaQA, clean and with the passage corrupted, and TriviaQA closed-book (fine partition, unexplored mass carried; $M{=}5$, $250$ prompts per cell). Percentages except set size (visible meanings) and Unexpl.\ (mean over adapters of the mass the search did not reach); gold in set: the presented set contains the gold meaning; Acc (rep.): accuracy of the representative meaning ($\arg\max\Plow$).}
\label{tab:csc}
\resizebox{\textwidth}{!}{
\begin{tabular}{lccccccc}
\toprule
Setting & $M$ & Commit & Acc$\mid$commit~$\uparrow$ & Gold in set~$\uparrow$ & Set size~$\downarrow$ & Acc (rep.)~$\uparrow$ & Unexpl. \\
\midrule
\multicolumn{8}{l}{\emph{Llama-3.1-8B}} \\
\rowcolor{gray!15} CoQA, clean & 5 & 40.4 & 95 & 87 & 2.52 & 75 & 0.51 \\
\rowcolor{gray!15} CoQA, corrupted context & 5 & 3.6 & 22 & 24 & 3.59 & 14 & 0.88 \\
\rowcolor{gray!15} TriviaQA, clean & 5 & 1.2 & 100 & 91 & 3.78 & 86 & 0.80 \\
\rowcolor{gray!15} TriviaQA, corrupted context & 5 & 0.0 & -- & 76 & 4.47 & 67 & 0.96 \\
\rowcolor{gray!15} TriviaQA closed-book & 5 & 32.0 & 96 & 88 & 2.80 & 77 & 0.58 \\
\multicolumn{8}{l}{\emph{Gemma-2-9B}} \\
\rowcolor{gray!15} CoQA, clean & 5 & 47.6 & 96 & 90 & 2.65 & 78 & 0.42 \\
\rowcolor{gray!15} CoQA, corrupted context & 5 & 3.6 & 33 & 28 & 3.30 & 12 & 0.83 \\
\rowcolor{gray!15} TriviaQA, clean & 5 & 1.2 & 100 & 88 & 2.51 & 83 & 0.78 \\
\rowcolor{gray!15} TriviaQA, corrupted context & 5 & 0.0 & -- & 76 & 3.14 & 66 & 0.86 \\
\rowcolor{gray!15} TriviaQA closed-book & 5 & 45.6 & 96 & 86 & 2.92 & 76 & 0.44 \\
\multicolumn{8}{l}{\emph{Qwen2.5-7B}} \\
\rowcolor{gray!15} CoQA, clean & 5 & 30.8 & 91 & 89 & 5.94 & 74 & 0.36 \\
\rowcolor{gray!15} CoQA, corrupted context & 5 & 7.6 & 32 & 26 & 9.38 & 13 & 0.73 \\
\rowcolor{gray!15} TriviaQA, clean & 5 & 6.4 & 94 & 86 & 6.44 & 76 & 0.79 \\
\rowcolor{gray!15} TriviaQA, corrupted context & 5 & 0.8 & 100 & 73 & 9.36 & 56 & 0.90 \\
\rowcolor{gray!15} TriviaQA closed-book & 5 & 24.4 & 95 & 76 & 7.16 & 56 & 0.56 \\
\bottomrule
\end{tabular}}

\end{table}

\paragraph{Token-space commitment under corrupted context.}
\Cref{tab:corrupt_commit} reports the token-space commit rates on clean and corrupted prompts discussed in \S\ref{sec:exp:results}~(iii).

\begin{figure}[h]\centering\includegraphics[width=\textwidth]{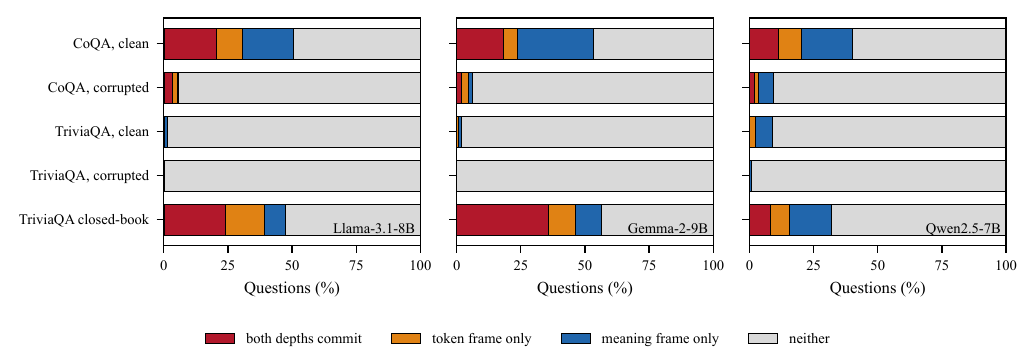}
\caption{Agreement between the two depths, per cell (the data of \cref{tab:agreement}).}
\label{fig:agreement}
\end{figure}

In \cref{tab:corrupt_commit}, token-space commitment on CoQA falls from $0.20$--$0.30$ on clean prompts to $0.036$--$0.052$ on corrupted ones, and the size of the token-space set alone separates corrupted from clean CoQA prompts with AUROC $0.78$--$0.83$. On TriviaQA with the passage the token space almost never closes, even on clean prompts (at most $2.4\%$), which is why the semantic space is needed there. \Cref{tab:agreement} and \cref{fig:agreement} split every question by which depths commit. On CoQA and closed-book TriviaQA both depths commit on $8$--$36\%$ of questions, the token space alone on $6$--$15\%$ and the semantic space alone on $8$--$30\%$; under corrupted context neither commits on $91$--$100\%$ of questions. \Cref{tab:depths} compares the two depths as detectors of the model's own wrong answers: on free-form answers the semantic space is ahead in all nine cells, by $2.6$ to $24.7$ points, and on the four-option datasets the two coincide.

\begin{table*}[h]
\centering
\setlength{\tabcolsep}{5pt}
\caption{\textbf{Token-space commitment under corrupted context} ($250$ clean and $250$ corrupted prompts per CoQA and TriviaQA cell, and the $500$ paired OpenBookQA questions of \cref{tab:ctc_mcq_full}; space = whole vocabulary, top-$100$ tokens by upper bound plus the remaining mass). A prompt is corrupted when its evidential passage is replaced or perturbed. OpenBookQA rows use the first-token caches of \cref{tab:ctc_mcq_full}; CoQA and TriviaQA rows use the credal-decoding caches of \cref{tab:csc} ($M{=}5$). The last column is the corrupted-versus-clean AUROC of the set size.}
\label{tab:corrupt_commit}
\resizebox{0.85\textwidth}{!}{
\begin{tabular}{llcccc}
\toprule
\textbf{Dataset} & \textbf{Backbone} & Commit rate, clean & Commit rate, corrupted & Median $|\Dset_1|$ clean / corrupted & AUROC of $|\Dset_1|$ \\
\midrule
\rowcolor{gray!15} OpenBookQA & Gemma-2-9B   & 0.898 & 0.706 & 1 / 1  & 0.881 \\
\rowcolor{gray!15} OpenBookQA & Llama-3.1-8B & 0.868 & 0.652 & 1 / 1  & 0.732 \\
\rowcolor{gray!15} OpenBookQA & Qwen2.5-7B   & 0.916 & 0.670 & 1 / 1  & 0.805 \\
\rowcolor{gray!15} CoQA & Gemma-2-9B   & 0.236 & 0.044 & 2 / 13 & 0.830 \\
\rowcolor{gray!15} CoQA & Llama-3.1-8B & 0.304 & 0.052 & 2 / 10 & 0.832 \\
\rowcolor{gray!15} CoQA & Qwen2.5-7B   & 0.204 & 0.036 & 3 / 20 & 0.777 \\
\rowcolor{gray!15} TriviaQA & Gemma-2-9B   & 0.008 & 0.000 & 75 / $>$1000 & 0.729 \\
\rowcolor{gray!15} TriviaQA & Llama-3.1-8B & 0.000 & 0.000 & 415 / $>$1000 & 0.629 \\
\rowcolor{gray!15} TriviaQA & Qwen2.5-7B   & 0.024 & 0.000 & 89 / $>$1000 & 0.731 \\
\bottomrule
\end{tabular}
}\end{table*}

\begin{table}[h]\centering\small
\caption{Agreement between depths on CoQA and TriviaQA, with the passage and closed-book (percent of questions): both CTC and CSC commit; only the token space; only the semantic space; neither.}
\label{tab:agreement}
\resizebox{\textwidth}{!}{
\begin{tabular}{lcccc}
\toprule
Setting & Both commit & Token space only & Semantic space only & Neither \\
\midrule
\multicolumn{5}{l}{\emph{Llama-3.1-8B}} \\
\rowcolor{gray!15} CoQA, clean & 20 & 10 & 20 & 50 \\
\rowcolor{gray!15} CoQA, corrupted context & 3 & 2 & 0 & 94 \\
\rowcolor{gray!15} TriviaQA, clean & 0 & 0 & 1 & 99 \\
\rowcolor{gray!15} TriviaQA, corrupted context & 0 & 0 & 0 & 100 \\
\rowcolor{gray!15} TriviaQA closed-book & 24 & 15 & 8 & 53 \\
\multicolumn{5}{l}{\emph{Gemma-2-9B}} \\
\rowcolor{gray!15} CoQA, clean & 18 & 6 & 30 & 47 \\
\rowcolor{gray!15} CoQA, corrupted context & 2 & 2 & 2 & 94 \\
\rowcolor{gray!15} TriviaQA, clean & 0 & 1 & 1 & 98 \\
\rowcolor{gray!15} TriviaQA, corrupted context & 0 & 0 & 0 & 100 \\
\rowcolor{gray!15} TriviaQA closed-book & 36 & 11 & 10 & 44 \\
\multicolumn{5}{l}{\emph{Qwen2.5-7B}} \\
\rowcolor{gray!15} CoQA, clean & 11 & 9 & 20 & 60 \\
\rowcolor{gray!15} CoQA, corrupted context & 2 & 2 & 6 & 91 \\
\rowcolor{gray!15} TriviaQA, clean & 0 & 2 & 6 & 91 \\
\rowcolor{gray!15} TriviaQA, corrupted context & 0 & 0 & 1 & 99 \\
\rowcolor{gray!15} TriviaQA closed-book & 8 & 8 & 16 & 68 \\
\bottomrule
\end{tabular}}

\end{table}

\begin{table*}[!h]
\centering
\setlength{\tabcolsep}{3pt}
\caption{\textbf{The same rule at both depths, on every dataset} (AUROC of each depth's score against the model's own answer being wrong, higher is better; CoQA, TriviaQA with and without the passage, OpenBookQA and ARC-Challenge). The main tables report one CLLM row, whose depth is fixed by the answer type rather than chosen per cell: the semantic space on free-form answers, the token space on the two four-option datasets, where a one-letter answer is its own meaning and the two depths are the same quantity. This table is the check on that rule. On free-form answers the semantic space is ahead in all nine cells, by $3$ to $25$ points. \textbf{Bold} = better of the two depths per column; grey rows are ours.}
\label{tab:depths}
\resizebox{\textwidth}{!}{\begin{tabular}{ll ccc ccc ccc ccc ccc}
\toprule
& & \multicolumn{3}{c}{\textbf{CoQA}} & \multicolumn{3}{c}{\textbf{TriviaQA (passage)}} & \multicolumn{3}{c}{\textbf{TriviaQA (closed book)}} & \multicolumn{3}{c}{\textbf{OpenBookQA}} & \multicolumn{3}{c}{\textbf{ARC-Challenge}} \\
\cmidrule(lr){3-5} \cmidrule(lr){6-8} \cmidrule(lr){9-11} \cmidrule(lr){12-14} \cmidrule(lr){15-17}
Depth (AUROC~$\uparrow$) & Candidates & Llama & Gemma & Qwen2.5 & Llama & Gemma & Qwen2.5 & Llama & Gemma & Qwen2.5 & Llama & Gemma & Qwen2.5 & Llama & Gemma & Qwen2.5 \\
\midrule
\rowcolor{gray!15} \textbf{CLLM (CTC)} & token space, first answer token & 0.754 & 0.752 & 0.714 & 0.711 & 0.744 & 0.755 & 0.764 & 0.706 & 0.536 & \textbf{0.917} & \textbf{0.942} & \textbf{0.917} & \textbf{0.781} & \textbf{0.830} & \textbf{0.807} \\
\rowcolor{gray!15} \textbf{CLLM (CSC)} & semantic space, meanings & \textbf{0.830} & \textbf{0.844} & \textbf{0.740} & \textbf{0.822} & \textbf{0.850} & \textbf{0.867} & \textbf{0.833} & \textbf{0.833} & \textbf{0.783} & \textbf{0.917} & \textbf{0.942} & \textbf{0.917} & \textbf{0.781} & \textbf{0.830} & \textbf{0.807} \\
\bottomrule
\end{tabular}
}
\end{table*}

\subsection{Ablations}
\label{app:ablations}
\textbf{Why the ablations are on a development split.} The expansion rule, the number of adapters and the meaning grouping are choices inside the method, so they are varied on a frozen CoQA development split that is disjoint from every test set, and the test cells are never used to select them. The adapter sweep uses nested subsets, so $M{=}2$ is a subset of $M{=}3$ and so on, which makes the comparison a statement about ensemble size rather than about which adapters happened to be drawn.

\subsubsection{Number of Adapters}
\label{app:ablation:m}
\begin{figure}[h]\centering\includegraphics[width=\textwidth]{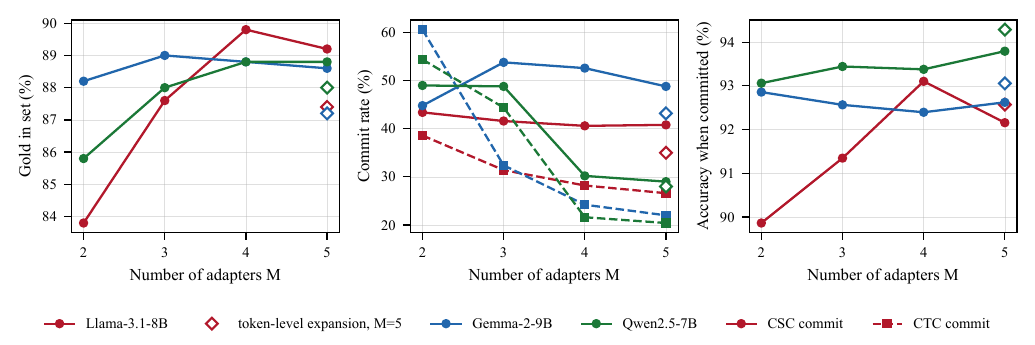}
\caption{\textbf{Number of adapters and expansion rule} (CoQA development split, $500$ prompts, nested subsets $M{=}2,3,4,5$; hollow marker: token-level expansion at $M{=}5$). Token-space commitment falls with $M$ on all three backbones while gold-in-set rises or holds.}
\label{fig:m_sweep}
\end{figure}

\cref{prop:monotone} implies that the commitment sets grow with $M$. \Cref{tab:m_sweep} runs the full pipeline on the CoQA development split ($500$ prompts, Llama) with nested subsets of the same adapters. Gold-in-set and accuracy when committed rise and commitment falls from $M{=}2$ to $4$ and are flat from $4$ to $5$; the token-space commit rate falls monotonically ($39$, $31$, $28$, $27\%$) as the intervals widen. \Cref{fig:m_sweep} runs the same sweep on all three backbones: token-space commitment falls with $M$ on every backbone, and gold-in-set rises from $M{=}2$ to $M{=}4$ on Llama and Qwen and stays within one point on Gemma.

\begin{table}[h]\centering\small
\caption{Number of adapters (CoQA development split, $500$ prompts, Llama-3.1-8B, sequence-level expansion, fine partition). Percentages except set size (visible meanings) and unexplored mass.}
\label{tab:m_sweep}
\begin{tabular}{lcccccc}
\toprule
$M$ & CTC commit & CSC commit & Acc$\mid$commit & Gold in set & Set size & Unexpl. \\
\midrule
\rowcolor{gray!15} 2 & 38.6 & 43.4 & 90 & 84 & 1.79 & 0.50 \\
\rowcolor{gray!15} 3 & 31.4 & 41.6 & 91 & 88 & 2.08 & 0.50 \\
\rowcolor{gray!15} 4 & 28.2 & 40.6 & 93 & 90 & 2.33 & 0.51 \\
\rowcolor{gray!15} 5 & 26.6 & 40.8 & 92 & 89 & 2.34 & 0.51 \\
\bottomrule
\end{tabular}
\end{table}

\subsubsection{Expansion Rule, Width and Unexplored Mass}
\label{app:ablation:expansion}
\emph{Expansion rule.} On the CoQA development split with the same five adapters, sequence-level expansion (prune a partial sequence when its upper mass falls below the best completed lower mass, \S\ref{sec:method}) against token-level expansion (keep only undominated next tokens at every step): CSC commit $40.8$ against $35.0\%$, accuracy when committed $92$ against $93\%$, gold in set $89$ against $87\%$, visible meanings $2.34$ against $2.52$, unexplored mass $0.51$ against $0.55$, completed sequences per prompt $10.9$ against $14.9$, decoding steps per prompt $72$ against $84$. Following the decision keeps fewer, more probable sequences and reaches more of the licensed mass. \emph{Width.} With three adapters and token-level expansion on the same split, gold was among the candidates on $82.6$, $84.0$ and $85.2\%$ of prompts at $B{=}4$, $8$ and $16$, with the width reached on $62$, $47$ and $36\%$ of prompts and $36$, $70$ and $129$ decoding steps per prompt; $B{=}8$ was fixed from this sweep before any test data was decoded. With five adapters and sequence-level expansion the width is reached on every prompt, since the top-$B$ partial sequences by upper mass are always kept, and the mass beyond them is carried as $u_m$. \emph{Carrying the unexplored mass.} Reading the same clean CoQA test prompts without $u_m$ in the upper bound and without the second commitment condition, that is treating the written-out completions as the whole sample space, raises the commit rate from $40.4$, $47.6$ and $30.8\%$ to $66.0$, $60.8$ and $38.0\%$ on Llama, Gemma and Qwen and lowers accuracy when committed from $95$, $96$ and $91\%$ to $90$, $91$ and $88\%$: the mass the search never reached is doing real work in the decision. \Cref{tab:expansion} repeats the expansion comparison on all three backbones: expanding by the decision raises CSC commitment on every backbone, by $5.8$, $5.6$ and $1.0$ points on Llama, Gemma and Qwen, and gold-in-set by $1.8$, $1.4$ and $0.8$ points, with accuracy when committed within $0.5$ points of token-level expansion.

\begin{table}[h]
\centering\footnotesize
\setlength{\tabcolsep}{5pt}
\caption{\textbf{Expansion rule} (CoQA development split, $500$ prompts, $M{=}5$, $B{=}8$). \emph{Decision}: prune a partial sequence when its upper mass falls below the best completed lower mass. \emph{Token}: keep only undominated next tokens. Commitment and gold-in-set improve everywhere; the step count falls on two backbones of three.}
\label{tab:expansion}
\resizebox{\textwidth}{!}{
\begin{tabular}{llccccc}
\toprule
Backbone & Expansion & CSC commit & Acc.\ when committed & Gold in set & Steps/prompt & Completed/prompt \\
\midrule
Llama-3.1-8B & decision & 40.8 & 92.2 & 89.2 & 72.4 & 10.9 \\
Llama-3.1-8B & token & 35.0 & 92.6 & 87.4 & 83.7 & 14.9 \\
Gemma-2-9B & decision & 48.8 & 92.6 & 88.6 & 73.0 & 23.9 \\
Gemma-2-9B & token & 43.2 & 93.1 & 87.2 & 65.1 & 27.2 \\
Qwen2.5-7B & decision & 29.0 & 93.8 & 88.8 & 97.5 & 14.9 \\
Qwen2.5-7B & token & 28.0 & 94.3 & 88.0 & 106.2 & 17.4 \\
\bottomrule
\end{tabular}}
\end{table}

\subsubsection{Meaning Grouping}
\label{app:ablation:meaning}
Grouping the same CoQA development candidates (Llama-3.1-8B, $M{=}5$, $500$ prompts) with an embedding model (BAAI/bge-base-en-v1.5, cosine) gives $1.46$ meanings per prompt at threshold $0.5$ and $3.22$ at $0.8$, with a gold-matching and a non-matching candidate in the same meaning on $66\%$ and $31\%$ of prompts; the adapters' own equivalence judgement gives $2.83$ meanings at $38\%$ without a threshold, and inspection of a random sample of $40$ of the flagged prompts finds $31$ to be answers the single gold string does not match (``Five'' against ``5'', ``He was.'' against ``Yes'', ``in excess of 3,500'' against ``over 3,500'') and $7$ to be genuine merges of answers at different granularity (``a friend of his'' with ``a man named Stephens''), the remaining two being too long to classify. The ``possibly equivalent'' partition (any adapter says yes) merges far more aggressively ($1.32$ meanings at $62\%$ impurity) and is reported only as a diagnostic.

\paragraph{The two meaning partitions.}
\cref{sec:method:depths} defines two partitions of the candidates: the \emph{surely} equivalent one, where every adapter judges a pair alike, which is what every reported number uses, and the \emph{possibly} equivalent one, where a single adapter suffices. \cref{tab:partition} contrasts them on the clean test prompts. The possible partition collapses the candidates into one or two meanings on Llama and Gemma and into fewer meanings on Qwen ($3.94$ and $1.78$), and the price is purity: the fraction of prompts on which a meaning holds both a gold-matching and a non-matching candidate rises from $19$--$60\%$ to $40$--$88\%$, reaching $88\%$ on Llama with TriviaQA. The gap between the two columns is the imprecision in meaning itself for this ensemble, and it is why the decision uses the sure partition: a rule that commits on a meaning built from one adapter's judgement would be committing to a class its own members do not agree on.

\begin{table}[h]
\centering\footnotesize
\setlength{\tabcolsep}{6pt}
\caption{\textbf{Sure against possible meaning partitions} on clean test prompts ($250$ per cell). Meanings: the mean number of meanings per prompt. Impurity: the percentage of prompts on which one meaning contains both a gold-matching and a non-matching candidate. Every result in the paper uses the sure partition.}
\label{tab:partition}
\begin{tabular}{llcccc}
\toprule
& & \multicolumn{2}{c}{Meanings per prompt} & \multicolumn{2}{c}{Impurity} \\
\cmidrule(lr){3-4} \cmidrule(lr){5-6}
Dataset & Backbone & sure & possible & sure & possible \\
\midrule
CoQA & Llama-3.1-8B & 2.98 & 1.41 & 30\% & 62\% \\
CoQA & Gemma-2-9B & 3.32 & 1.42 & 30\% & 64\% \\
CoQA & Qwen2.5-7B & 6.88 & 3.94 & 19\% & 40\% \\
TriviaQA & Llama-3.1-8B & 3.80 & 1.01 & 60\% & 88\% \\
TriviaQA & Gemma-2-9B & 2.52 & 1.00 & 38\% & 66\% \\
TriviaQA & Qwen2.5-7B & 6.62 & 1.78 & 33\% & 62\% \\
\bottomrule
\end{tabular}
\end{table}

\subsection{Cost-Sensitive Commitment}
\label{app:ablation:cost_sensitive}
\textcolor{blue}{The cost-sensitive rule is the $\Gamma$-maximin decision of imprecise probability~\citep{troffaes07}: choose the action whose lower expected utility is largest. With utility $1$ for a correct answer, $-\kappa$ for a wrong one and $0$ for abstaining, committing to $a^*$ has lower expected utility $\min_{q\in\mathcal{P}}\bigl[q(a^*)-\kappa\,(1-q(a^*))\bigr]=\Plow(a^*)(1+\kappa)-\kappa$, so it beats abstaining exactly when $\Plow(a^*)\geq\kappa/(1+\kappa)$.} \cref{tab:cost} sweeps the declared cost $\kappa$ of \S\ref{sec:method:depths}, where the model commits to the representative meaning when $\Plow(c^*)\geq\kappa/(1+\kappa)$. Raising $\kappa$ trades coverage for accuracy: on CoQA, Llama answers $54\%$ of questions at $93\%$ accuracy when a wrong answer costs half a right one, and $16\%$ at $100\%$ accuracy when it costs four times as much. The more interesting column is the first. The rule as stated, which has no $\kappa$ and no threshold, lands on the same operating point as the cost rule at $\kappa=1$, and not merely in aggregate: across the nine cells the two agree on $98.4$ to $100\%$ of individual prompts, and in the $9$ prompts of $2250$ where they differ it is always the rule as stated committing where the cost rule would not. The parameter-free rule therefore behaves as if a wrong answer costs exactly what a right answer is worth, which is the neutral choice a system makes when it has not been told the costs, and $\kappa$ is available when a deployment knows better.

\begin{table}[h]
\centering\footnotesize
\setlength{\tabcolsep}{4pt}
\caption{\textbf{Cost-sensitive commitment} on clean CoQA and TriviaQA prompts ($250$ per cell). Cov.: fraction of questions answered; Acc.: accuracy of the committed answer. ``Rule as stated'' is the parameter-free rule of \cref{eq:dominance} with the unexplored-mass condition; the remaining columns commit when $\Plow(c^*)\geq\kappa/(1+\kappa)$. A dash marks a column that commits on no question in that cell.}
\label{tab:cost}
\resizebox{\textwidth}{!}{\begin{tabular}{llcccccccccc}
\toprule
& & \multicolumn{2}{c}{Rule as stated} & \multicolumn{2}{c}{$\kappa=0.5$} & \multicolumn{2}{c}{$\kappa=1$} & \multicolumn{2}{c}{$\kappa=2$} & \multicolumn{2}{c}{$\kappa=4$} \\
\cmidrule(lr){3-4} \cmidrule(lr){5-6} \cmidrule(lr){7-8} \cmidrule(lr){9-10} \cmidrule(lr){11-12}
Dataset & Backbone & Cov. & Acc.~$\uparrow$ & Cov. & Acc.~$\uparrow$ & Cov. & Acc.~$\uparrow$ & Cov. & Acc.~$\uparrow$ & Cov. & Acc.~$\uparrow$ \\
\midrule
CoQA & Llama-3.1-8B & 0.40 & 0.95 & 0.54 & 0.93 & 0.40 & 0.95 & 0.28 & 0.97 & 0.16 & 1.00 \\
CoQA & Gemma-2-9B & 0.48 & 0.96 & 0.56 & 0.94 & 0.47 & 0.96 & 0.31 & 0.99 & 0.18 & 1.00 \\
CoQA & Qwen2.5-7B & 0.31 & 0.91 & 0.37 & 0.90 & 0.31 & 0.91 & 0.22 & 0.98 & 0.14 & 1.00 \\
TriviaQA & Llama-3.1-8B & 0.01 & 1.00 & 0.07 & 1.00 & 0.01 & 1.00 & 0.00 & -- & 0.00 & -- \\
TriviaQA & Gemma-2-9B & 0.01 & 1.00 & 0.17 & 0.98 & 0.01 & 1.00 & 0.00 & -- & 0.00 & -- \\
TriviaQA & Qwen2.5-7B & 0.06 & 0.94 & 0.20 & 0.98 & 0.06 & 0.94 & 0.00 & 1.00 & 0.00 & -- \\
TriviaQA closed-book & Llama-3.1-8B & 0.32 & 0.96 & 0.44 & 0.94 & 0.32 & 0.96 & 0.23 & 0.98 & 0.17 & 1.00 \\
TriviaQA closed-book & Gemma-2-9B & 0.46 & 0.96 & 0.50 & 0.96 & 0.45 & 0.97 & 0.36 & 0.97 & 0.26 & 0.98 \\
TriviaQA closed-book & Qwen2.5-7B & 0.24 & 0.95 & 0.29 & 0.94 & 0.23 & 0.95 & 0.15 & 0.95 & 0.11 & 0.96 \\
\bottomrule
\end{tabular}
}
\end{table}

\subsection{Computational Cost}
\label{app:cost}
Measured on the runs behind \cref{tab:csc} ($M{=}5$, $B{=}8$, $L{=}32$, KV-cached decoding, adapters swapped in one process). Credal decoding (which yields CTC, the candidate set and every adapter's sequence probabilities in one pass) takes $7$--$12$ s per prompt on CoQA and closed-book TriviaQA, $11$--$18$ s on TriviaQA with the passage and $3$--$5$ s on the eight-token OpenBookQA setting; the equivalence judgements add $5$--$12$ s per prompt on CoQA and closed-book TriviaQA and $8$--$24$ s on TriviaQA with the passage, where the candidate set is largest. Times are for Llama-3.1-8B on one L4 (24\,GB) and Gemma-2-9B on one A100 (40\,GB) and are similar on the two. Peak memory is $21$\,GB for Llama, $32$\,GB for Gemma and $16$\,GB for Qwen2.5-7B, dominated by the base model and the KV caches of $B$ partial sequences. Semantic entropy for a single model ($K{=}16$ samples, $L{=}32$) takes $2$--$3$ s per prompt on the same hardware, and the mean-pooled ensemble costs exactly the decoding pass above with the same adapters. Decoding steps per prompt are $16$--$25$ on OpenBookQA, $68$--$89$ on CoQA and closed-book TriviaQA and $135$--$217$ on TriviaQA with the passage, and forward passes per prompt are $M$ times these. Adapter training is a one-off cost of about one GPU-hour per adapter and is not part of inference.

\section{Broader Impact}
\label{app:impact}

This work concerns reliability tooling for instruction-tuned Large Language Models, specifically a commitment rule that answers only when no plausible fine-tuning of the model would prefer a different answer, and otherwise returns the set of answers that remain plausible. The most likely positive consequence of CLLM-style commitment is reducing fluent-but-wrong outputs in deployment-critical settings such as healthcare, legal, scientific, and educational assistants, where confidently-wrong answers carry larger downstream cost than abstentions. The credal-set view exposes the spread of the credal set rather than collapsing it; reviewers, regulators, and downstream users can read off both \emph{what} the model predicts and \emph{how stable} the prediction is across plausible adapters.

We see two material risks. First, a committed answer can be misread as a correct one: commitment means that every plausible fine-tuning in the realised ensemble agrees, not that the agreement is right. Inputs on which all adapters share a mistake (for example poisoned context they all absorbed the same way) are committed with high confidence, and no rule defined on the ensemble can detect a failure the whole ensemble shares. Reliance on CLLM as a sole correctness gate, without complementary retrieval or human review, would inherit this blind spot.

Second, an ensemble of $M{=}5$ adapters raises both training and inference cost relative to a single fine-tune; \S\ref{app:cost} reports the overhead. A forward pass with a LoRA adapter costs roughly $1.1\times$ a base forward pass, so the total inference cost scales with $M$. In carbon-conscious deployments the trade-off should be made explicit: CTC requires no generation and costs one $M$-adapter forward pass, while CSC additionally requires credal decoding and the equivalence judgements. We recommend reporting these costs alongside commit rate and accuracy when CLLM is integrated into a downstream system.

The protocols draw on existing public datasets (OpenBookQA, CoQA, TriviaQA, ARC-Challenge), and we do not collect or release new data.

\ifdefined\appendixstandalone
\bibliography{bibliography}
\bibliographystyle{iclr2027_conference}
\fi
\providecommand\appendixend{}
\appendixend

\end{document}